\documentclass{article}

\usepackage{arxiv}

\UseRawInputEncoding
\usepackage{amssymb}
\usepackage[utf8]{inputenc} 
\usepackage[T1]{fontenc}    
\usepackage{hyperref}       
\usepackage{url}            
\usepackage{booktabs}       
\usepackage{amsfonts}       
\usepackage{nicefrac}       
\usepackage{microtype}      
\usepackage{caption}
\usepackage{subcaption}
\usepackage{verbatim}
\usepackage{xcolor}
\usepackage{bbm}
\usepackage{amsthm}
\usepackage{bm}
\usepackage{graphicx}
\usepackage{pgf}
\usepackage{amsmath}
\usepackage{cleveref}
\usepackage{widebar}
\usepackage{mathtools}
\newtheorem{assumption}{Assumption}

\newtheorem{theorem}{Theorem}

\newtheorem{lemma}{Lemma}

\newcommand{\amse}[0]{\mathrm{AMSE}}
\newcommand{\cov}[0]{\mathrm{Cov}}
\newcommand{\norm}[2]{\left\lVert#1\right\rVert_{#2}}

\newcommand{\expect}[1]{\mathbb E\left[#1\right]}

\newcommand{\transpose}[1]{#1^\top}
\newcommand{\trace}[0]{\mathrm{Tr}}

\DeclareCaptionLabelFormat{andtable}{#1~#2  \&  \tablename~\thetable}

\title{The Mean-Squared Error of Double Q-Learning}

\author{Wentao Weng \\
Tsinghua University \\
\texttt{wwt17@mails.tsinghua.edu.cn} \\
\And
Harsh Gupta \\
University of Illinois at Urbana-Champaign\\
\texttt{hgupta10@illinois.edu} \\
\And
Niao\ He \\
University of Illinois at Urbana-Champaign\\
\texttt{niaohe@illinois.edu} \\
\And
Lei\ Ying \\
University of Michigan, Ann Arbor \\
\texttt{leiying@umich.edu} \\
\And
R.\ Srikant \\
University of Illinois at Urbana-Champaign\\
\texttt{rsrikant@illinois.edu} \\
}

\begin{document}

\maketitle

\begin{abstract}

In this paper, we establish a theoretical comparison between the asymptotic mean-squared error of Double Q-learning and Q-learning. Our result builds upon an analysis for linear stochastic approximation based on Lyapunov equations and applies to both tabular setting and with linear function approximation, provided that the optimal policy is unique and the algorithms converge. We show that the asymptotic mean-squared error of Double Q-learning is exactly equal to that of Q-learning if Double Q-learning uses twice the learning rate of Q-learning and outputs the average of its two estimators. We also present some practical implications of this theoretical observation using simulations.

\end{abstract}







\section{Introduction}
Reinforcement learning (RL) seeks to design efficient algorithms to find optimal policies for Markov Decision Processes (MDPs) without any knowledge of the underlying model (known as model-free learning) \cite{Sutton_18}. In this paper, we study the performance of double Q-learning \cite{Hasselt_10,Hasselt_16}, which is a popular variant of the standard Watkins's model-free Q-learning algorithm \cite{Watkins_92,Watkins_1989}. Double Q-learning was proposed to remedy the stability issues associated with the standard Q-learning algorithm (due to maximization bias of the Q-function) by using two estimators instead of one. It has been shown empirically that double Q-learning finds a better policy in the tabular setting~\cite{Hasselt_10} and converges faster when coupled with deep neural networks for function approximation \cite{Hasselt_16}. Several variations of Double Q-learning were proposed in \cite{zhang_2017, anschel2017averaged}. However, to the best of our knowledge, there has been no analysis of double Q-learning vis-\`a-vis how it performs theoretically as compared to standard Q-learning. The objective of this paper is to address this question by providing a tight theoretical comparison between double Q-learning and Q-learning while also drawing experimental insights that allow us to corroborate the theory.

Stochastic Approximation (SA) has proven to be a powerful framework to analyze reinforcement learning algorithms \cite{Borkar_2009, Benveniste_12, Kushner_2003}. Several different types of guarantees for various reinforcement learning algorithms have been established using techniques from stochastic approximation. The most commonplace result is the asymptotic convergence of algorithms by analyzing the stability of an associated ODE. Examples include \cite{TsiVan_1997},  \cite{Sutton_1988} for classical TD-learning with linear function approximation, \cite{BorkarMeyn_00} for synchronous Q-learning, \cite{LeeHe_2019} for double TD-learning, and \cite{Melo_2008, Lee_Dong_He_2019} for Q-learning with linear function approximation. To the best of our knowledge, establishing the convergence of double Q-learning with linear function approximation remains an open problem \cite{LeeHe_2019}. 
Although establishing asymptotic convergence of an algorithm is a useful theoretical goal, quantifying the finite-time convergence rate of an algorithm can be more useful in providing actionable insight to practitioners. There has been a significant body of recent work in this context. Finite-time analyses of TD-learning with either decaying or constant learning rate can be found in \cite{Srikant_2019, gupta_2019, dalal_2018, dalal_18, laksh_18, bhandari_2018}. Finite-time error bounds for synchronous Q-learning can be found in \cite{Chen_2019, Chen_2020} and  for asynchronous Q-learning in \cite{Qu_2020}. 
This line of work primarily focuses on providing upper bounds on the error, thereby failing to make a tight comparison between a pair of algorithms  designed for solving the same problem. Recently, several papers developed tight error bounds for SA and RL algorithms, including \cite{Devraj_2017,Devraj_2020, chenDevAdiMeyn_20, Hu_2019}.  

In this paper, we focus on comparing Double Q-learning with standard Q-learning, both theoretically and experimentally. We observe that through a particular linearization technique introduced in \cite{Devraj_2017}, both Double Q-learning and Q-learning can be formulated as instances of Linear Stochastic Approximation (LSA). 
We further utilize a recent result \cite{chenDevAdiMeyn_20} that characterizes the asymptotic variance of an LSA recursion by a Lyapunov equation. By analyzing these associated Lyapunov equations for both Q-learning and Double Q-learning, we establish bounds comparing these two algorithms.

The main contributions of this work are two-fold:

\textbf{(1) Theoretical Contributions:}
We consider asynchronous Double Q-learning and Q-learning with linear function approximation with decaying step-size rules (as special cases of the more general LSA paradigm). Under the assumptions that the optimal policy is unique, both the algorithms converge and the step-size for Double Q-learning is twice that of Q-learning, we show that  the asymptotic mean-squared errors of the two estimators of Double Q-learning are strictly worse than that of the estimator in Q-learning, while the asymptotic mean-squared error of the average of the Double Q-learning estimators is indeed equal to that of the Q-learning estimator. This result brings interesting practical insight, leading to our second set of contributions.

\textbf{(2) Experimental Insights:} Combining results from our experiments and previous work, we have the following observations:
\begin{enumerate}
    \item If Double Q-learning and Q-learning use the same step-size rule, Q-learning has a faster rate of convergence initially but suffers from a higher mean-squared error. This phenomenon is observed both in our simulations and in earlier work on variants of Double TD-learning \cite{LeeHe_2019}. 
    
    \item If the step-size used for Double Q-learning is twice that of Q-learning, then Double Q-learning achieves faster initial convergence rate, at the cost of a possibly worse mean-squared error than Q-learning. However, if the final output is the average of the two estimators in Double Q-learning, then its asymptotic mean-squared error is the same as that of Q-learning. 
    
\end{enumerate}
The thumb rule that these observations suggest is that one should use a higher learning rate for Double Q-learning while using the average of its two estimators as the output. 

\section{Q-learning and Double Q-learning}

Consider a Markov Decision Process (MDP) specified by $\left(\mathcal{S},\mathcal{A},P,R,\gamma\right)$. Here $\mathcal{S}$ is the finite state space, $\mathcal{A}$ is the finite action space, $P \in \mathbb{R}^{|\mathcal{S}||\mathcal{A}|\times |\mathcal{S}|}$ is the action-dependent transition matrix,  $R \in \mathbb{R}^{|\mathcal{S}|\times |\mathcal{A}|}$ is the reward matrix, and  $\gamma \in [0,1)$ is the discount factor. Upon selecting an action $a$ at state $s$, the agent will transit to the next state $s'$ with probability $P((s,a),s')$ and receive an immediate reward $R(s,a)$. 

A policy is a mapping from a state to an action, which specifies the action to be taken at each state. It is well known that the optimal policy can be obtained by solving the so-called Bellman equation \cite{Bertsekas_96,Sutton_18} for the state-action value function, also called the Q-function:
\begin{equation}\label{eq:def-Q-function}
Q^*(s,a) = R(s,a)+\gamma \sum_{s' \in \mathcal{S}} P((s,a),s')\max_{a'\in \mathcal A}Q^*(s',a').
\end{equation}

In reinforcement learning, the goal is to estimate the Q-function from samples, without knowing the parameters of the underlying MDP. For simplicity, we assume the MDP is operated under a fixed behavioral policy, and we observe a sample trajectory of the induced Markov chain $\{(S_1,A_1),\cdots,(S_n,A_n),\cdots\}$. Let $X_n = (S_n,A_n)$ and define $\mathcal{X}=\mathcal{S}\times\mathcal{A}.$  Since the state space could be fairly large, function approximation is typically used to approximate the $Q$-function. In this work, we focus on linear function approximation for its tractability. The goal is to find an optimal estimator $\theta^* \in \mathbb{R}^d$, such that $Q^* \approx \transpose{\Phi}\theta^*$,
where $\Phi = (\phi(s^1,a^1),\cdots,\phi(s^{|\mathcal{X}|},a^{|\mathcal{X}}|)) \in \mathbb{R}^{d \times |\mathcal{X}|}$, and $\phi(s,a) \in \mathbb{R}^d$ are given feature vectors associated with pairs of states and actions.

\subsection{Q-learning}

We first consider asynchronous Q-learning \cite{Watkins_92,Watkins_1989} with linear function approximation. Let $\Phi = (\phi(x_1),\cdots,\phi(x_{|\mathcal{X}|})) \in \mathbb{R}^{d\times |\mathcal{X}|}$ be the matrix consisting of columns of feature vectors. We let $\pi_{\theta}$ denote the greedy policy with respect to the parameter vector $\theta,$ i.e., $\pi_\theta(s)=\arg\max_a  \phi(s,a)^T\theta,$ where we assume that we break ties in the maximization according to some known rule. 
For ease of notation, we define $H(\theta_1,\theta_2,s): = \transpose{\phi(s,\pi_{\theta_1}(s))}\theta_2$. This function estimates the Q-function based on $\theta_2$ while the action is selected from the greedy policy given by $\theta_1$. When observations on the sample path proceed to $(X_n,S_{n+1})$, Q-learning updates the parameter $\theta$ according to the equation: 
\begin{equation}\label{eq:Q-update}
\theta_{n+1}=\theta_n+\alpha_{n} \phi(X_n)\left(R(X_n)+\gamma H(\theta_n,\theta_n,S_{n+1}) -\transpose{\phi(X_n)}\theta_n\right),
\end{equation}
where $\alpha_n$ is an appropriately chosen step-size, also known as the learning rate.

\subsection{Double Q-learning}

To improve the performance of Q-learning, Double Q-learning was introduced in \cite{Hasselt_10,Hasselt_16}. We consider the Double Q-learning with linear function approximation here. Double Q-learning maintains two estimators $\theta_n^A, \theta_n^B$, which are updated to estimate $Q^*$ based on the sample path $\{X_n\}$ in the following manner:
\begin{equation}\label{eq:doubleQ-update}
\begin{aligned}
\theta_{n+1}^A &=\theta_n^A+\beta_n\delta_{n}\left( \phi(X_n)\left(R(X_n)+\gamma H(\theta_n^A,\theta_n^B,S_{n+1}) -\transpose{\phi(X_n)}\theta_n^A\right)\right), \\
\theta_{n+1}^B &=\theta_n^B+(1-\beta_n)\delta_{n}\left(\phi(X_n)\left(R(X_n)+\gamma H(\theta_n^B,\theta_n^A,S_{n+1}) -\transpose{\phi(X_n)}\theta_n^B\right)\right),
\end{aligned}
\end{equation}
where $\beta_n$ are IID Bernoulli random variables equal to one w.p. $1/2$ and $\delta_n$ is the step-size. Note that at each time instant, only one of $\theta_A$ or $\theta_B$ is updated.

\subsection{Linear Stochastic Approximation}\label{sec:linearization}

Under the assumptions that the optimal policy is unique, the ordinary differential equation (ODE) associated with Q-learning is stable and other technical assumptions, it has been argued in \cite{Devraj_2017} that the asymptotic variance of $Q$-learning can be studied by considering the recursion
\begin{equation}\label{eq:Q-update linear}
\theta_{n+1}=\theta_n+\alpha_{n} \phi(X_n)\left(R(X_n)+\gamma \transpose{\phi(S_{n+1}, \pi^*(S_{n+1}))}\theta_n -\transpose{\phi(X_n)}\theta_n\right),
\end{equation}
where $\pi^*$ is the optimal policy $\pi_{\theta^*}$ based on $\theta^*$. Here and throughout, as in \cite{Devraj_2017}, we assume that the Q-learning and Double Q-learning algorithms converge to some $\theta^*.$  We refer the reader to \cite{Devraj_2017} for details.

Using a similar argument, one can show that the asymptotic variance of Double Q-learning can be studied by considering the following recursion:
\begin{equation}\label{eq:doubleQ-update linear}
\begin{aligned}
\theta_{n+1}^A &=\theta_n^A+\beta_n\delta_{n}\left( \phi(X_n)\left(R(X_n)+\gamma \transpose{\phi(S_{n+1}, \pi^*(S_{n+1}))}\theta_n^B -\transpose{\phi(X_n)}\theta_n^A\right)\right), \\
\theta_{n+1}^B &=\theta_n^B+(1-\beta_n)\delta_{n}\left(\phi(X_n)\left(R(X_n)+\gamma \transpose{\phi(S_{n+1}, \pi^*(S_{n+1}))}\theta_n^A -\transpose{\phi(X_n)}\theta_n^B\right)\right).
\end{aligned}
\end{equation}

Our comparison of the asymptotic mean-squared errors of Q-learning and Double Q-learning will use (\ref{eq:Q-update linear})-(\ref{eq:doubleQ-update linear}).  In practice, however, one is typically interested in how quickly one learns the optimal policy which cannot be measured very well using the mean-squared error metric. Later, we will see that our simulations indicate that the insights we obtain from mean-squared error analysis hold even for learning the optimal policy.

\section{Main Results}

In this section, we present our main results. Before we do, we first review the results on asymptotic variance of linear stochastic approximation in \cite{chenDevAdiMeyn_20} and use these to compare the asymptotic variances of Q-learning and Double Q-learning.

\subsection{Preliminaries}

Consider the linear stochastic approximation recursion:
\begin{equation}\label{eq: LSA}
    \xi_{n+1}=\xi_n+\frac{g}{n} \left(A(Y_n)\xi_n+b(Y_n)\right),
\end{equation}
where $g$ is a positive constant, $Y_n$ is an irreducible, aperiodic Markov Chain on a finite state space, $A$ and $b$ are a random matrix and a random vector, respectively, which are determined by $Y_n.$ Without loss of generality, we assume $\xi_n$ converges to $\xi^* = 0$. If $\xi^* \not = 0$, we can subtract $\xi^*$ from $\xi_n$. Define the asymptotic covariance of $\xi_n$ to be
$$\Sigma_\infty=\lim_{n\rightarrow\infty} n \expect{\xi_n\xi_n^T}.$$ The following result is from \cite{chenDevAdiMeyn_20}.
\begin{theorem}\label{thm: LSA}
Suppose that $\bar{A}:=\expect{A(Y_\infty)}$, and $\frac{1}{2}I+g\bar{A}$ is a Hurwitz matrix, i.e., its eigenvalues have negative real parts, and $\Sigma_b:=\expect{b(Y_1)\transpose{b(Y_1)}}+\sum_{n=2}^{\infty} \expect{b(Y_n)\transpose{b(Y_1)} + b(Y_1)\transpose{b(Y_n)}},$\footnote{In \cite{chenDevAdiMeyn_20}, the asymptotic covariance $\Sigma_b$ is defined by $\sum_{n=-\infty}^{+\infty} \expect{b(Y_n)\transpose{b(Y_0)}}.$ Since in our setting the time starts from 1, we equivalently write the covariance as $\expect{b(Y_1)\transpose{b(Y_1)}}+\sum_{n=2}^{+\infty} \expect{b(Y_n)\transpose{b(Y_1)} + b(Y_1)\transpose{b(Y_n)}}.$ An earlier version of the paper had an incorrect equation, and we thank Tobias Sutter for noticing this.} where $Y_\infty$ is notation for a random variable with the same distribution as the stationary distribution of the Markov chain $\{Y_n\}.$ Then, $\Sigma_\infty$ is the unique solution to the Lyapunov equation
\begin{equation}\label{eq: Lyapunov}
    \Sigma_\infty \left(\frac{1}{2}I+g\transpose{\bar{A}}\right)+ \left(\frac{1}{2}I+g\bar{A}\right)\Sigma_\infty + g^2\Sigma_b=0.
\end{equation}
\end{theorem}

In the next subsection, we use the above result to establish the relationship between the asymptotic covariances of Q-learning and Double Q-learning.

\subsection{Comparison of Q-learning and Double Q-learning}

Throughout this section, we assume that $\theta^*=0$ without loss of generality. If $\theta^*\neq 0,$ the results can hold by subtracting $\theta^*$ from the estimators of Q-learning and Double Q-learning. Our main result is stated in the following theorem.

\begin{theorem}\label{thm: comparison}
Define the asymptotic mean-squared error of Q-learning to be 
$$\amse(\theta):=\lim_{n\rightarrow\infty} n \expect{\theta_n^T\theta_n},$$  
the asymptotic mean-squared error of the estimator in Double Q-learning to be 
$$\amse(\theta^A):=\lim_{n\rightarrow\infty} n \expect{\transpose{(\theta_n^A)}\theta_n^A},$$ 
and the asymptotic mean-squared error of the average of the two Double Q-learning estimators to be
$$\amse\left(\frac{\theta^A+\theta^B}{2}\right)=\lim_{n\rightarrow\infty}\frac{1}{4} n \expect{\transpose{(\theta_n^A+\theta_n^B)}(\theta_n^A+\theta_n^B)}.$$
Let the step sizes of Q-learning and Double Q-learning be $\alpha_n=g/n$ and $\delta_n=2g/n$, where $g$ is a positive constant. 
Then there exists some $g_0>0$, such that for any $g> g_0$,  the following results hold:
\begin{enumerate}
    \item $\amse(\theta^A)\geq \amse(\theta)$, and
    \item $\amse(\frac{\theta^A+\theta^B}{2})=\amse(\theta).$
\end{enumerate}
\end{theorem}

Before we present the proof of the above result, we make some remarks.\\

\textbf{Remark 1.} The condition $g> g_0$ is tied to the sufficient conditions for stability of the ODEs associated with covariance equations of Q-learning and Double Q-learning~\cite{chenDevAdiMeyn_20}. If we consider both in tabular case, namely, $\Phi$ is exactly an identity matrix with dimension $|\mathcal{X}|$. Let $\mu_{\min}$ be the minimum probability of a state $x \in \mathcal{X}$ in the stationary distribution $\mu$. In this case, the  results hold so long as $g > \frac{1}{\mu_{\min}(1-\gamma)}$, which is a common assumption used in the analysis of tabular Q-learning \cite{Qu_2020}. 

\textbf{Remark 2.} As mentioned in the introduction to this paper, Double Q-learning can be slower initially due to the fact that only half the samples are used to estimate each of its estimators. One way to speed up the initial convergence rate is to double the learning rate. Our results here show that the asymptotic mean-squared error of Double Q-learning in that case will be at least as large as that of Q-learning; however, if the output of Double Q-learning is the average of its two estimators, the asymptotic mean-squared error is exactly equal to that of Q-learning with half the learning rate. Thus, Double Q-learning learns faster without sacrificing asymptotic mean-squared error. This suggests that increasing the learning rate of Double Q-learning while averaging the output can have significant benefits, which we verify using simulations in the next section. Now, we are ready to present the proof of the theorem.

\paragraph{Proof of Theorem~\ref{thm: comparison}:}

Recall from Section \ref{sec:linearization} that the asymptotic variance of Q-learning can be studied by considering the following recursion:

\begin{equation}\label{eq:Q-update linear proof}
\theta_{n+1}=\theta_n+\alpha_{n} \phi(X_n)\left(R(X_n)+\gamma \transpose{\phi(S_{n+1}, \pi^*(S_{n+1}))}\theta_n -\transpose{\phi(X_n)}\theta_n\right).
\end{equation}
Similarly, one can show that the asymptotic variance of double Q-learning can be studied by considering the following recursion:
\begin{equation}\label{eq:doubleQ-update linear proof}
\begin{aligned}
\theta_{n+1}^A &=\theta_n^A+\beta_n\delta_{n}\left( \phi(X_n)\left(R(X_n)+\gamma \transpose{\phi(S_{n+1}, \pi^*(S_{n+1}))}\theta_n^B -\transpose{\phi(X_n)}\theta_n^A\right)\right), \\
\theta_{n+1}^B &=\theta_n^B+(1-\beta_n)\delta_{n}\left(\phi(X_n)\left(R(X_n)+\gamma \transpose{\phi(S_{n+1}, \pi^*(S_{n+1}))}\theta_n^A -\transpose{\phi(X_n)}\theta_n^B\right)\right).
\end{aligned}
\end{equation}
For ease of notation, let $Z_n = (X_n,S_{n+1})$. It is shown in \cite{Chen_2019} that $\{Z_n\}$ is also an aperiodic and irreducible Markov chain. Let us define the following: $b(Z_n) = \phi(X_n)R(X_n)$, $A_1(Z_n) = \phi(X_n)\transpose{\phi(X_n)}$, $ A_2(Z_n) = \gamma\phi(X_n)\transpose{\phi(S_{n+1}, \pi^*(S_{n+1}))}, A(Z_n) = A_2(Z_n) - A_1(Z_n)$. Using these definitions, we can rewrite \eqref{eq:Q-update linear proof} and \eqref{eq:doubleQ-update linear proof} as:
\begin{equation}\label{eq:Q-update-linear-easy}
\theta_{n+1} = \theta_n +\alpha_n\left(b(Z_n)+A_2(Z_n)\theta_n - A_1(Z_n)\theta_n\right).
\end{equation}
and
\begin{equation}\label{eq:DoubleQ-update-linear-easy}
\begin{aligned}
\theta_{n+1}^A &=\theta_n^A+\beta_n\delta_{n}\left( b(Z_n)+A_2(Z_n)\theta_n^B -A_1(Z_n)\theta_n^A\right), \\
\theta_{n+1}^B &=\theta_n^B+(1-\beta_n)\delta_{n}\left(b(Z_n)+A_2(Z_n)\theta_n^A -A_1(Z_n)\theta_n^B\right),
\end{aligned}
\end{equation}
respectively. Let $U_n = \transpose{(\transpose{(\theta_n^A)},\transpose{(\theta_n^B)})}$. We can further write \eqref{eq:DoubleQ-update-linear-easy} in a more compact form as:
\begin{equation}\label{eq:doubleQ-final}
U_{n + 1}  = U_n + \alpha_n\bigg[\begin{pmatrix}
-2\beta_nA_1(Z_n) & 2\beta_nA_2(Z_n)\\
2(1 - \beta_n)A_2(Z_n) & -2(1 - \beta_n)A_1(Z_n)
\end{pmatrix}U_n + \begin{pmatrix}
2\beta_nb(Z_n)\\
2(1 - \beta_n)b(Z_n)
\end{pmatrix}\bigg].
\end{equation}

Let $\mu$ denote the steady-state probability vector for the Markov chain $\{X_n\}$. Let $D$ be a diagonal matrix of dimension $|\mathcal{X}|$ such that $D_{ii} = \mu_i$. We have $\widebar{A}_1 = \expect{A_1(Z_{\infty})} = \Phi D \transpose{\Phi}, \widebar{A}_2 = \expect{A_2(Z_{\infty})} = \gamma\Phi DPS_{\pi^*} \transpose{\Phi}$, where $S_{\pi^*}$ is the action selection matrix of the optimal policy $\pi^*$ such that $S_{\pi^*}(s,(s,\pi^*(s))) = 1$ for $s \in \mathcal{S}$. Denote $ \widebar{A} = \widebar{A}_2-\widebar{A}_1$.

We will now use Theorem \ref{thm: LSA} to prove our result. Let $\Sigma_{\infty}^Q = \lim_{n\rightarrow\infty} n\expect{\theta_n\theta_n^T}$ and $\Sigma_{\infty}^D = \lim_{n\rightarrow\infty} n\expect{U_nU_n^T}$. Clearly, $\amse(\theta) = \trace(\Sigma_{\infty}^Q)$. Applying Theorem \ref{thm: LSA} to \eqref{eq:Q-update-linear-easy} and \eqref{eq:doubleQ-final}:

\begin{equation}\label{eq:Q-LSA-result}
    \Sigma_\infty^Q \left(\frac{1}{2}I+g\transpose{\bar{A}}\right)+ \left(\frac{1}{2}I+g\bar{A}\right)\Sigma_\infty^Q + g^2(B_1 + B_2)=0,
\end{equation}
and
\begin{equation}\label{eq:doubleQ-LSA-result}
    \Sigma_\infty^D \left(\frac{1}{2}I+g\transpose{\widebar{A}_D}\right)+ \left(\frac{1}{2}I+g\widebar{A}_D\right)\Sigma_\infty^D + g^2 \Sigma_{b}^D =0,
\end{equation}
where $B_2 =  \frac{1}{2}\expect{\sum_{n=2}^{\infty} (b(X_1)\transpose{b(X_n)}+b(X_n)\transpose{b(X_1)})}$, $B_1 =\expect{b(X_1)\transpose{b(X_1)}} + B_2$, $\widebar{A}_D = \begin{pmatrix}
-\widebar{A}_1 & \widebar{A}_2\\
\widebar{A}_2 & -\widebar{A}_1 \end{pmatrix}$, and $\Sigma_b^D = 2\begin{pmatrix}
B_1 & B_2\\
B_2 & B_1 \end{pmatrix}$. Because of the symmetry in the two estimators comprising Double Q-learning, we observe that $\Sigma_{\infty}^D$ will have the following structure: $\Sigma_{\infty}^D = \begin{pmatrix}
V & C\\
C & V
\end{pmatrix}
$, where $$V=\lim_{n\rightarrow\infty}n\expect{\theta^A_n \transpose{(\theta^A_n)}}=\lim_{n\rightarrow\infty}n\expect{\theta^B_n \transpose{(\theta^B_n)}},\qquad C=\lim_{n\rightarrow\infty}n \expect{\theta^A_n \transpose{(\theta^B_n)}}.$$ Coupling this observation with \eqref{eq:doubleQ-LSA-result} yields
\begin{equation}\label{eq:analysis-equations}
    \begin{pmatrix}
V & C\\
C & V
\end{pmatrix} + g\begin{pmatrix}
V & C\\
C & V
\end{pmatrix}\begin{pmatrix}
-\widebar{A}_1 & \widebar{A}_2\\
\widebar{A}_2 & -\widebar{A}_1 \end{pmatrix}^T+ g\begin{pmatrix}
-\widebar{A}_1 & \widebar{A}_2\\
\widebar{A}_2 & -\widebar{A}_1 \end{pmatrix}\begin{pmatrix}
V & C\\
C & V
\end{pmatrix} + 2g^2\begin{pmatrix}
B_1 & B_2\\
B_2 & B_1 \end{pmatrix} =0.
\end{equation}
Summing the first two blocks (row-wise) of matrices in the above equation, we get
\begin{equation}\label{eq:analysis-equations-1}
V + C + g(V + C)(\widebar{A}_2 - \widebar{A}_1)^T + g(\widebar{A}_2 - \widebar{A}_1)(V + C) + 2g^2(B_1+B_2) = 0.
\end{equation}
Next, define $g_0:=\inf\{g\geq 0: g\max(\lambda_{\max}(\bar{A}),\lambda_{\max}(\bar{A}_D)) < -1\}$, where $\lambda_{\max}(A)$ denotes the real part of the maximum eigenvalue of $A$. Note that $g_0$ exists since both $\bar{A}$ and $\bar{A}_D$ are  Hurwitz, under the assumption that Q-learning and Double Q-learning both converge \cite{Chen_1998}. As a result, for any $g> g_0$,  $\frac{1}{2}I+g\bar{A}$ is Hurwitz. Therefore, the solution $V+C$ to the above equation and the solution $\Sigma_\infty$ to \eqref{eq:Q-LSA-result} are unique \cite{Chen_1998}. Similarly, we also note that the solution to \eqref{eq:analysis-equations} is also unique as $\frac{1}{2}I+g\bar{A}_D$ is Hurwitz whenever $g>g_0$. 

Comparing the above equation with \eqref{eq:Q-LSA-result}, we get $\Sigma_{\infty}^Q = \frac{V + C}{2}$.
Next, we observe that $\trace(V) \geq \trace(C)$. The reasoning behind that is as follows:
\begin{align*}
    \lim_{n \rightarrow\infty}n\expect{(\theta^A_{n} - \theta^B_{n})^T(\theta^A_{n} - \theta^B_{n}
    )} &\geq 0\\
    \Rightarrow 2\lim_{n \rightarrow\infty}n\big\{\expect{(\theta_{n}^A)^T\theta_{n}^A} - \expect{(\theta_{n}^B)^T\theta_{n}^A}\big\} &\geq 0
    \Rightarrow \trace(V) - \trace(C)\geq 0,
\end{align*}
where the second inequality follows from the symmetry in the two estimators comprising double Q-learning. Using $\trace(V) \geq \trace(C),$ we get
$
    \trace(V) \geq \trace\left(\frac{V + C}{2}\right) = \trace(\Sigma_{\infty}^Q).
$
This equation proves our first result. To prove the second result, we observe that
\begin{align*}
    \amse\left(\frac{\theta^A + \theta^B}{2}\right) = \frac{1}{2}\amse(\theta^A) + \frac{1}{2}\trace(C)= \frac{1}{2}(\trace(V) + \trace(C)) = \trace(\Sigma_{\infty}^Q).
\end{align*}
\hfill \qedsymbol

\section{Numerical Results}
In this section, we provide numerical comparisons between Double Q-learning and Q-learning on Baird's Example \cite{Baird_1995},  GridWorld \cite{geramifard_2013},
CartPole \cite{barto_1983} and an example of maximization bias from \cite{Sutton_18} \footnote{Codes are at  \url{https://github.com/wentaoweng/The-Mean-Squared-Error-of-Double-Q-Learning}.}. We investigate four algorithms: 1) Q-learning using step size $\alpha_n$, denoted as Q in plots; 2) Double Q-learning using step size $\alpha_n$, denoted as D-Q; 3) Double Q-learning using step size equal to $2\alpha_n$, denoted as D-Q with twice the step size; 4) Double Q-learning using step size equal to $2\alpha_n$ and returning the average estimator $({\theta_n^A+\theta_n^B})/{2}$, denoted as D-Q average with twice step size. For the vanilla Double Q-learning, we always use $\theta_n^A$ as its estimator. 

For the first two experiments, we plot the logarithm of the mean-squared error for each algorithm. We set the step size $\alpha_n = \frac{1000}{n+10000}$. The optimal estimator, $\theta^*$, is calculated by solving the projected Bellman equation \cite{Lee_Dong_He_2019} based on the Markov chain. Sample paths start in state $1$ in Baird's Example, and state $(1,1)$ in GridWorld. We use the uniformly random policy as the behavioral policy, i.e., each valid action is taken with equal probability in any given state. Initialization of $\theta_1,\theta_1^A,\theta_1^B$ are set the same and are uniformly sampled from $[0,2]^d$, where $d$ is the dimension of features.  Results in each plot reflect the average over $100$ sample paths.
\subsection{Baird's Example}
\begin{figure}
    \begin{subfigure}{0.5\textwidth}
    \centering
    \includegraphics[scale=0.32]{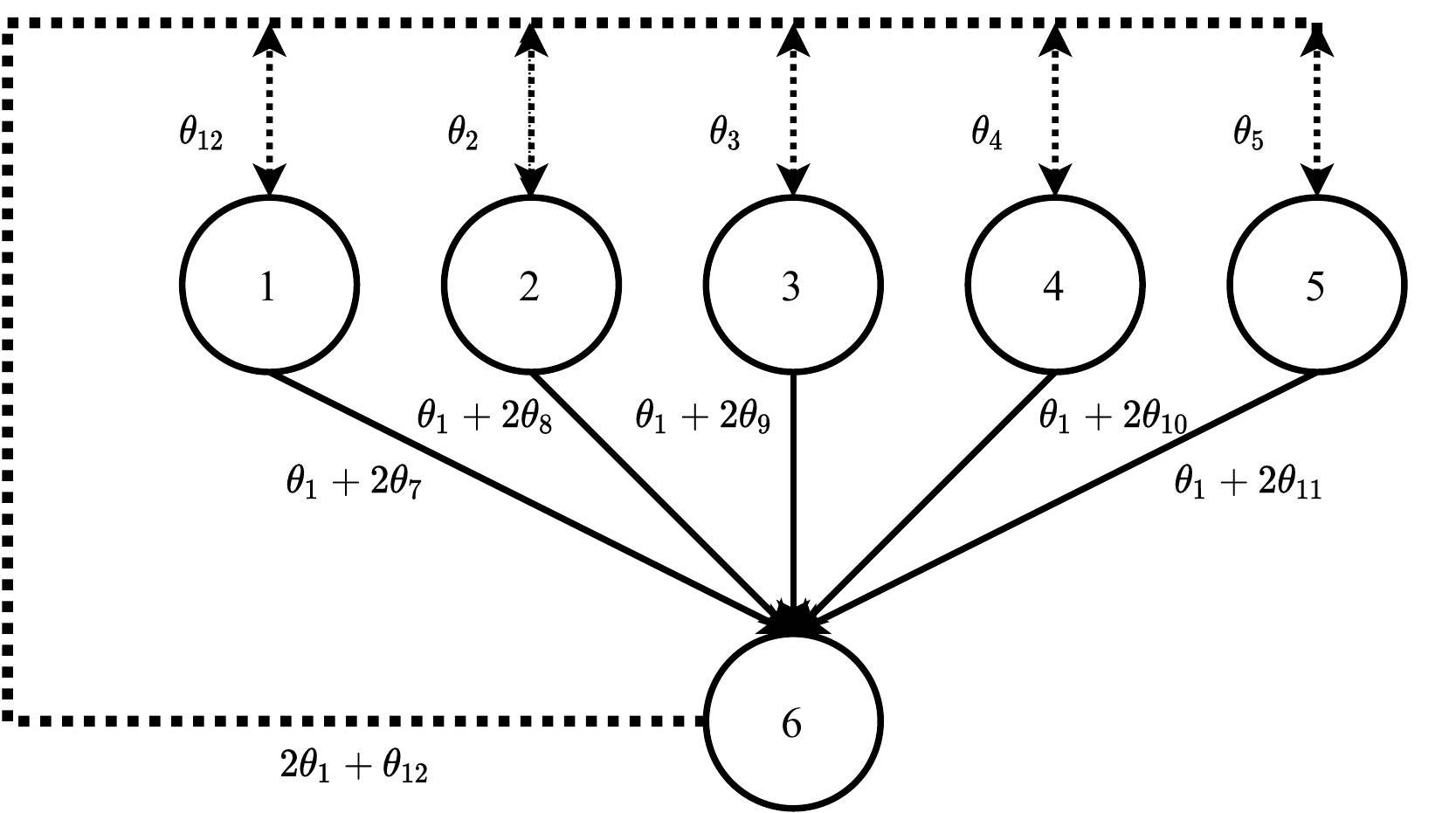}
    \caption{Baird's Example \cite{Baird_1995}}
    \label{fig:Baird-Example}
    \end{subfigure}
    \begin{subfigure}{0.5\textwidth}
        \centering
        \includegraphics[width=3in]{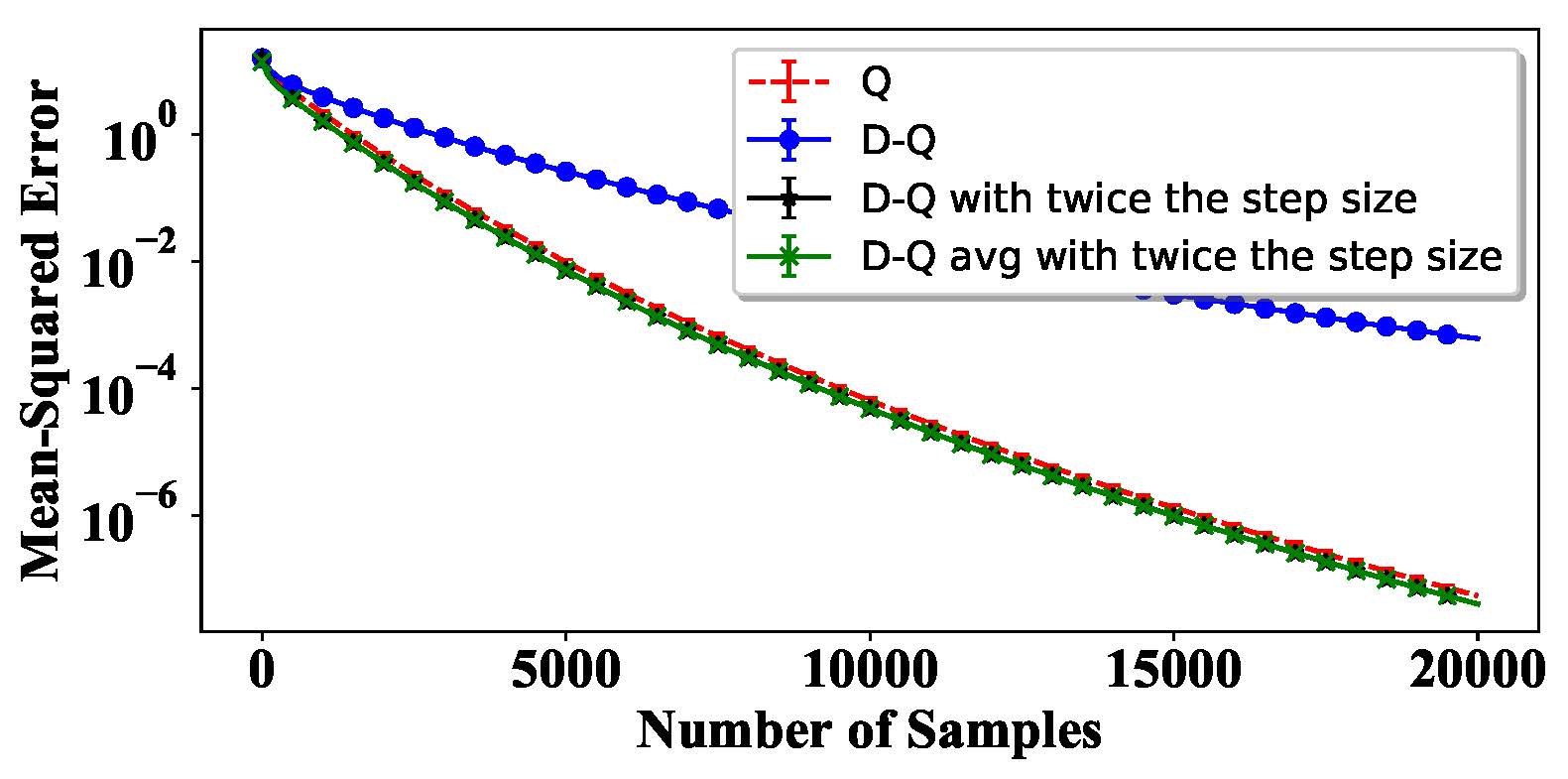}
        \caption{Zero Reward}
        \label{fig:Baird-type0}
    \end{subfigure}
    
\quad
    \begin{subfigure}{0.4\textwidth}
        \centering
        \includegraphics[width=3in]{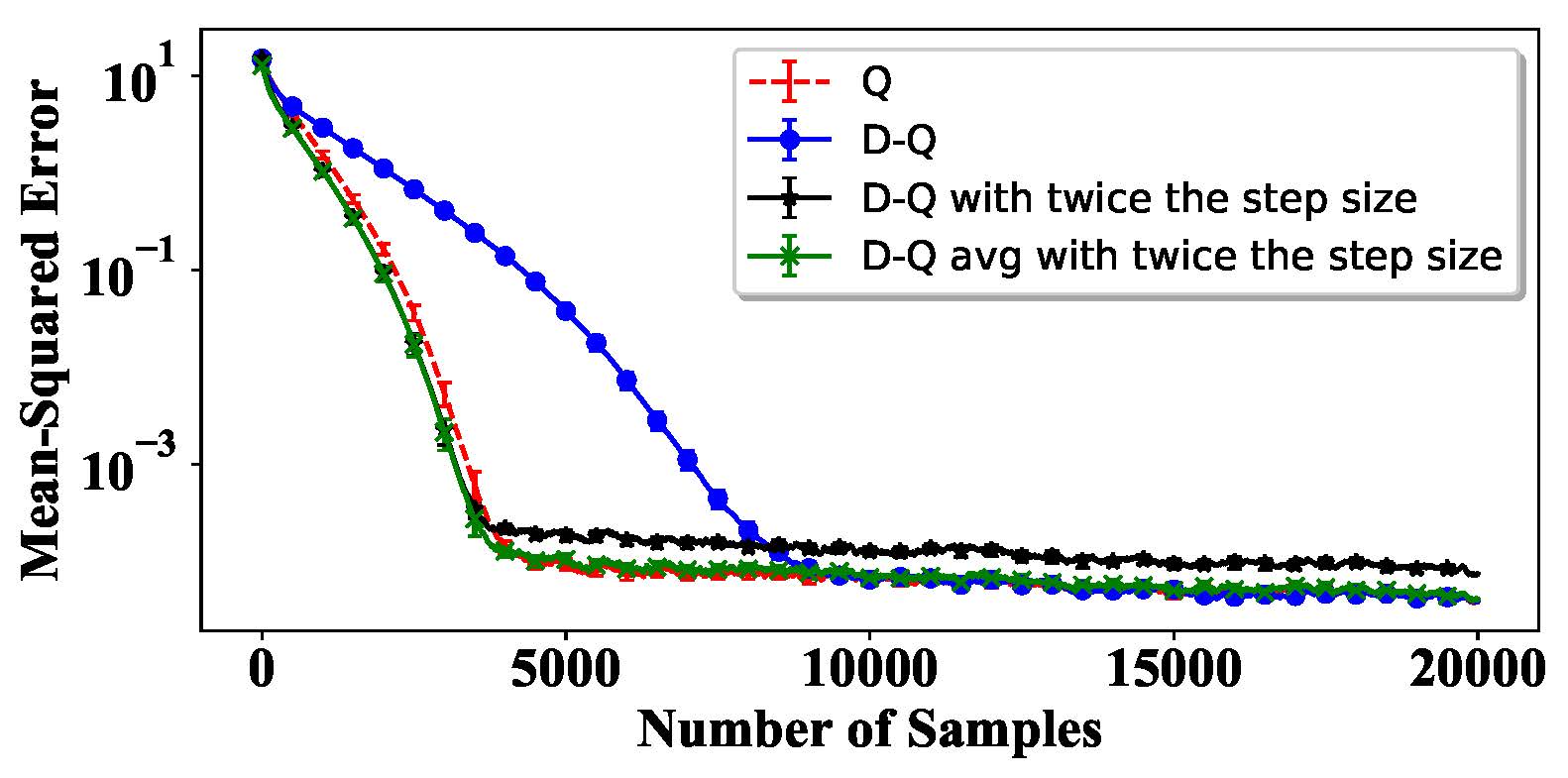}
        \caption{Small Random Reward}
        \label{fig:Baird-type1}
    \end{subfigure}
\quad
    \begin{subfigure}{0.6\textwidth}
        \centering
        \includegraphics[width=3in]{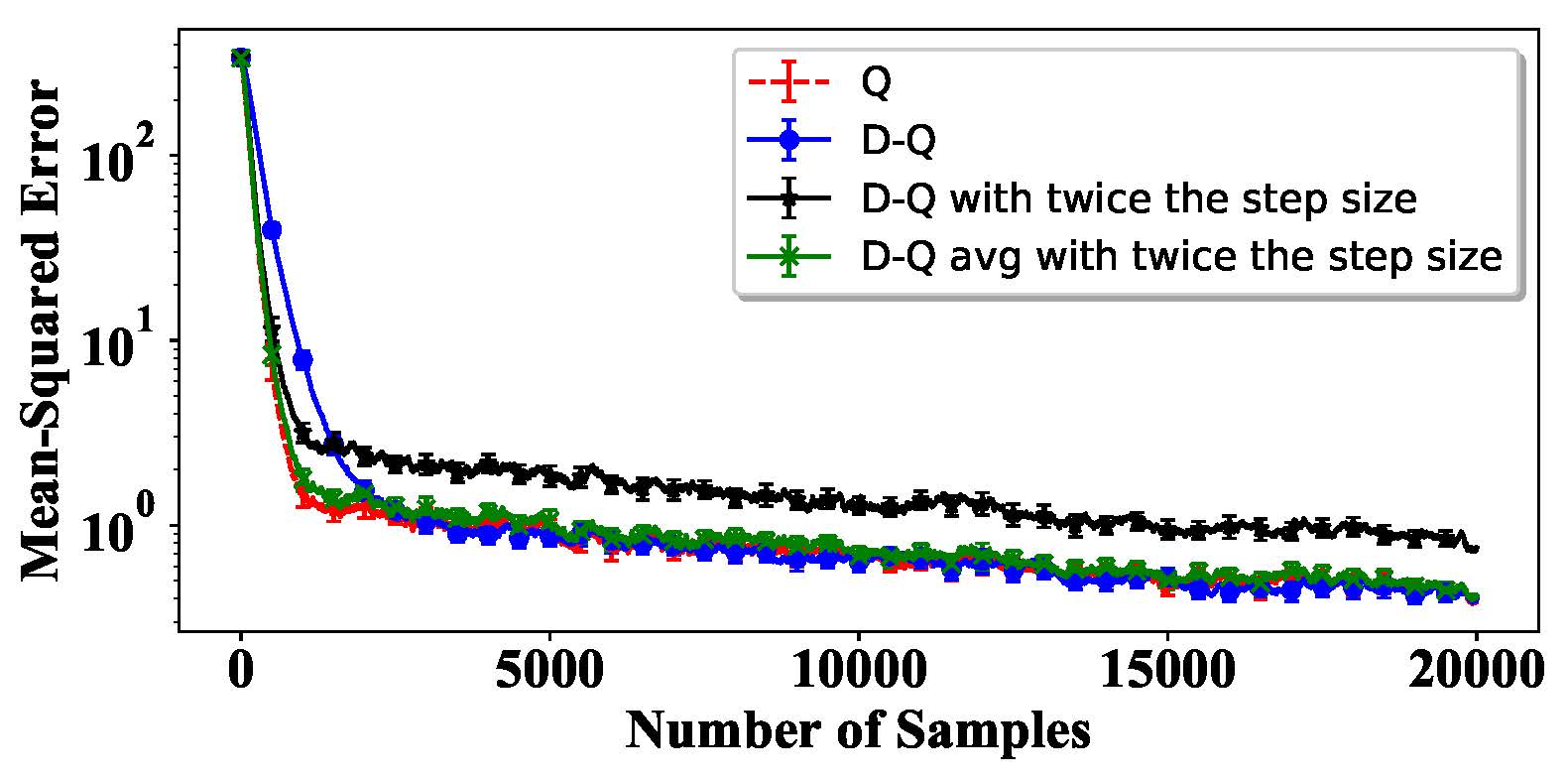}
        \caption{Large Random Reward}
        \label{fig:Baird-type2}
    \end{subfigure}
    \caption{Simulation results for Baird's example. The y-axis is in log scale.}
\end{figure}

The first environment we consider is the popular Baird's Example which was used to prove that Q-learning with linear function approximation may diverge~ \cite{Chen_2019,Baird_1995}. It is a simple Markov chain as shown in Fig. \ref{fig:Baird-Example} with $6$ states and $2$ actions (represented by the dotted line and the solid line respectively). When the action represented by the dotted line is taken, the agent transits to one of the first five states randomly. When an action represented by a solid line is taken, the agent transits to state $6$. The $Q$-function is approximated by a parameter $\theta \in \mathbb{R}^{12}$, where the specific linear combination is shown next to the corresponding action. For the reward function $R(s,a)$, $1 \leq s \leq 6, 1 \leq a \leq 2$, we explore different settings:
1) \textbf{Zero Reward:} the reward $R(s,a)$ is uniformly zero;
2) \textbf{Small Random Reward:} the reward $R(s,a)$ is sampled uniformly from $[-0.05,0.05]$; 
3) \textbf{Large Random Reward:} the reward $R(s,a)$ is sampled uniformly from $[-50,50]$. Our theory applies to Small Random Reward and Large Random Reward because the optimal policy is unique in these two cases, but simulations indicate that our insight works more generally even in the case of Zero Reward. Although Baird's example was originally proposed to make Q-learning diverge when $\gamma$ is large, we study the case $\gamma = 0.8$ where all algorithms converge. Results are presented in Fig. \ref{fig:Baird-type0}, \ref{fig:Baird-type1}, and \ref{fig:Baird-type2}. 

In all the three scenarios, we observe that Double Q-learning converges much slower than Q-learning at an early stage, when using the same step-size .
When using a step size $2\alpha_n$, we observe that Double Q-learning converges slightly faster than Q-learning in Fig. \ref{fig:Baird-type0}, Fig. \ref{fig:Baird-type1}, and almost at the same speed in Fig. \ref{fig:Baird-type2}. However, the mean-squared error is much worse than that of Q-learning as shown in Fig. \ref{fig:Baird-type1} and Fig. \ref{fig:Baird-type2}. Finally, by simply using the averaged estimator, Double Q-learning obtains both faster convergence rate and smaller mean-squared error, which matches with our theory. 

\subsection{GridWorld} \label{sec:gridworld}
\begin{figure}[!ht]
    \begin{subfigure}{0.5\textwidth}
    \centering
    \caption{An Example of $3\times 3$ GridWorld}
    \includegraphics[scale=0.12]{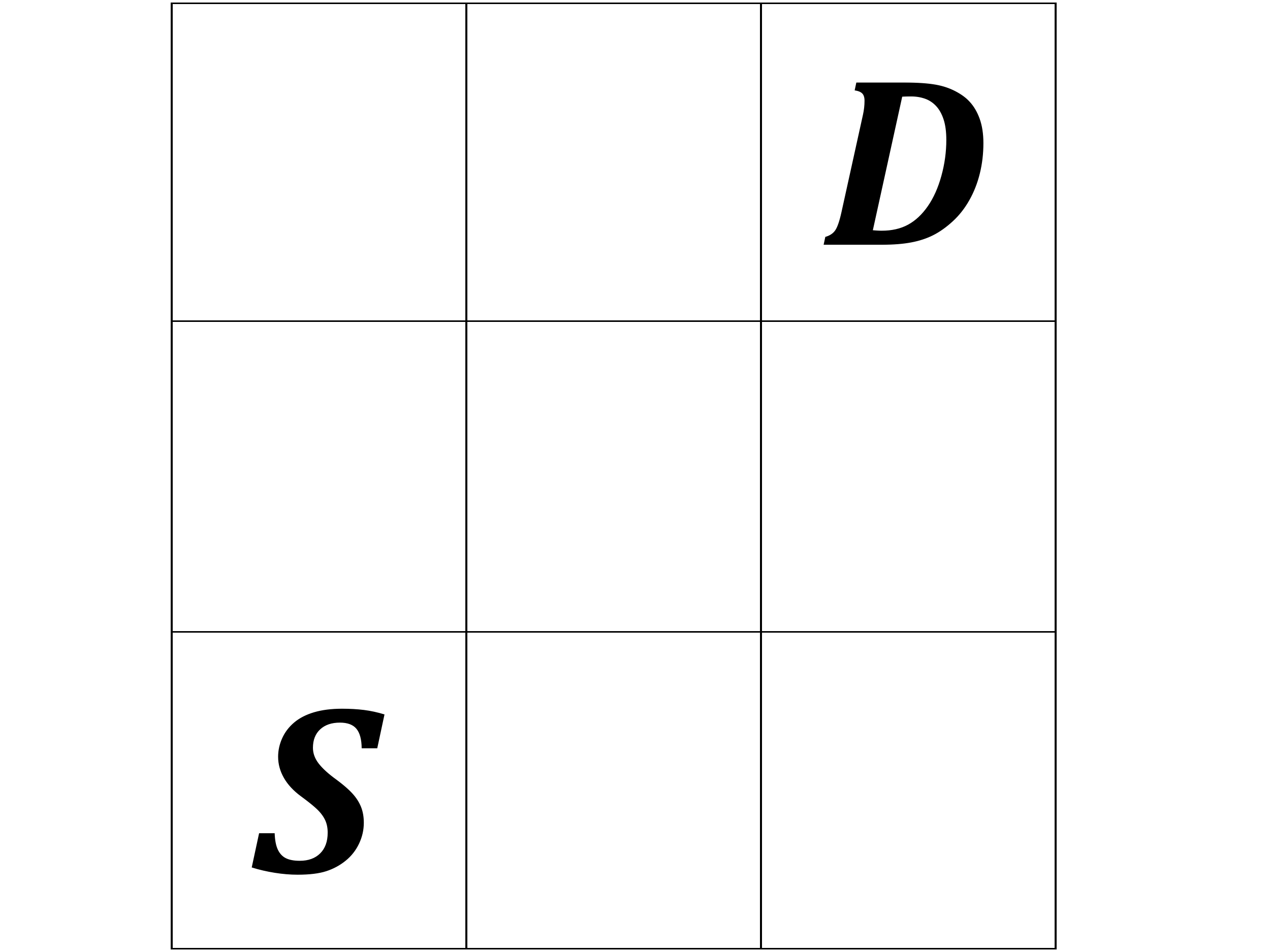}
    \label{fig:grid-Example}
    \end{subfigure}
    \begin{subfigure}{0.5\textwidth}
        \centering
        \caption{$3\times 3$ GridWorld}
        \includegraphics[width=2.5in]{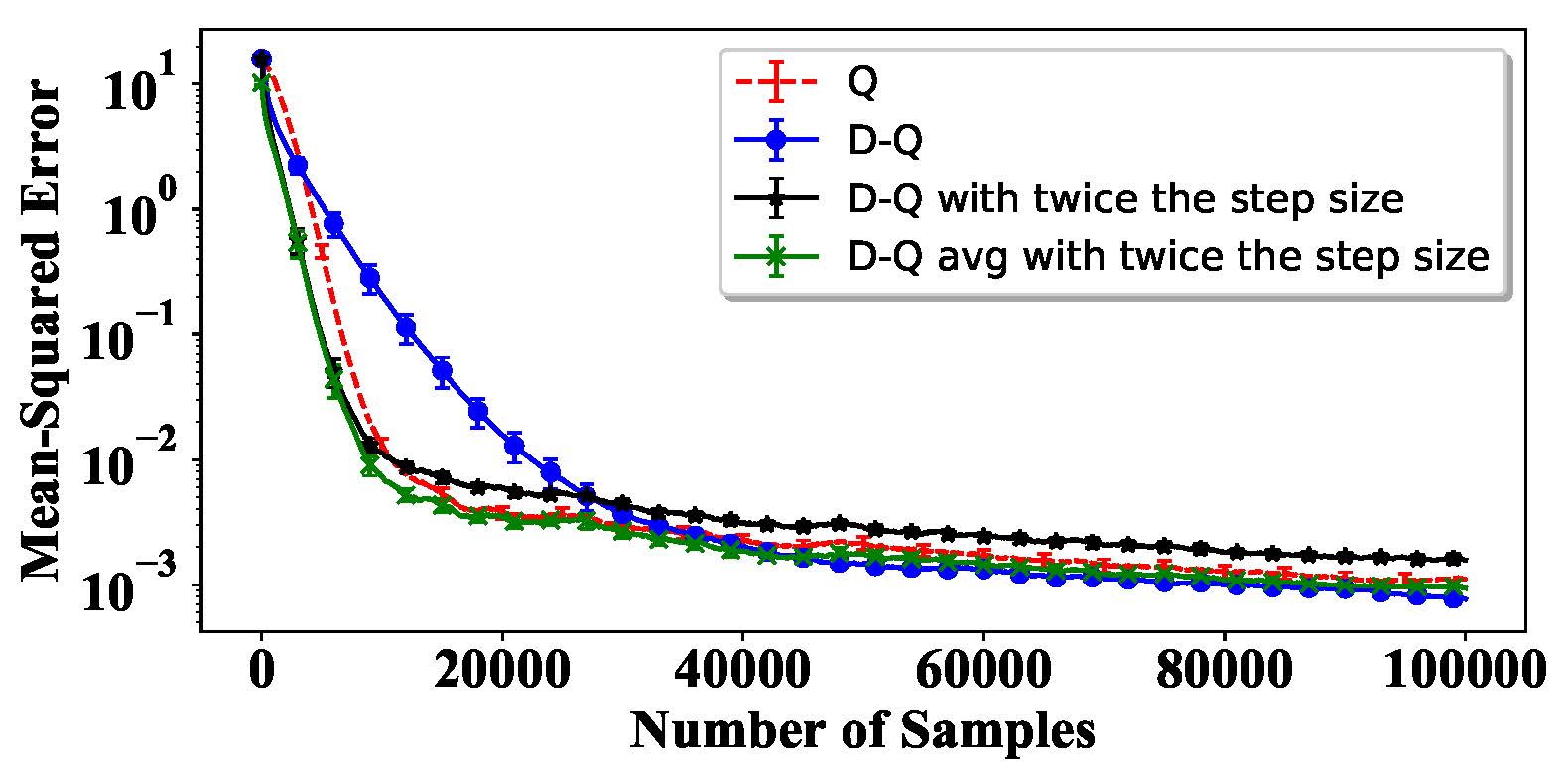}
        \label{fig:3x3-grid}
    \end{subfigure}
    
\quad
    \begin{subfigure}{0.4\textwidth}
        \centering
        \includegraphics[width=2.5in]{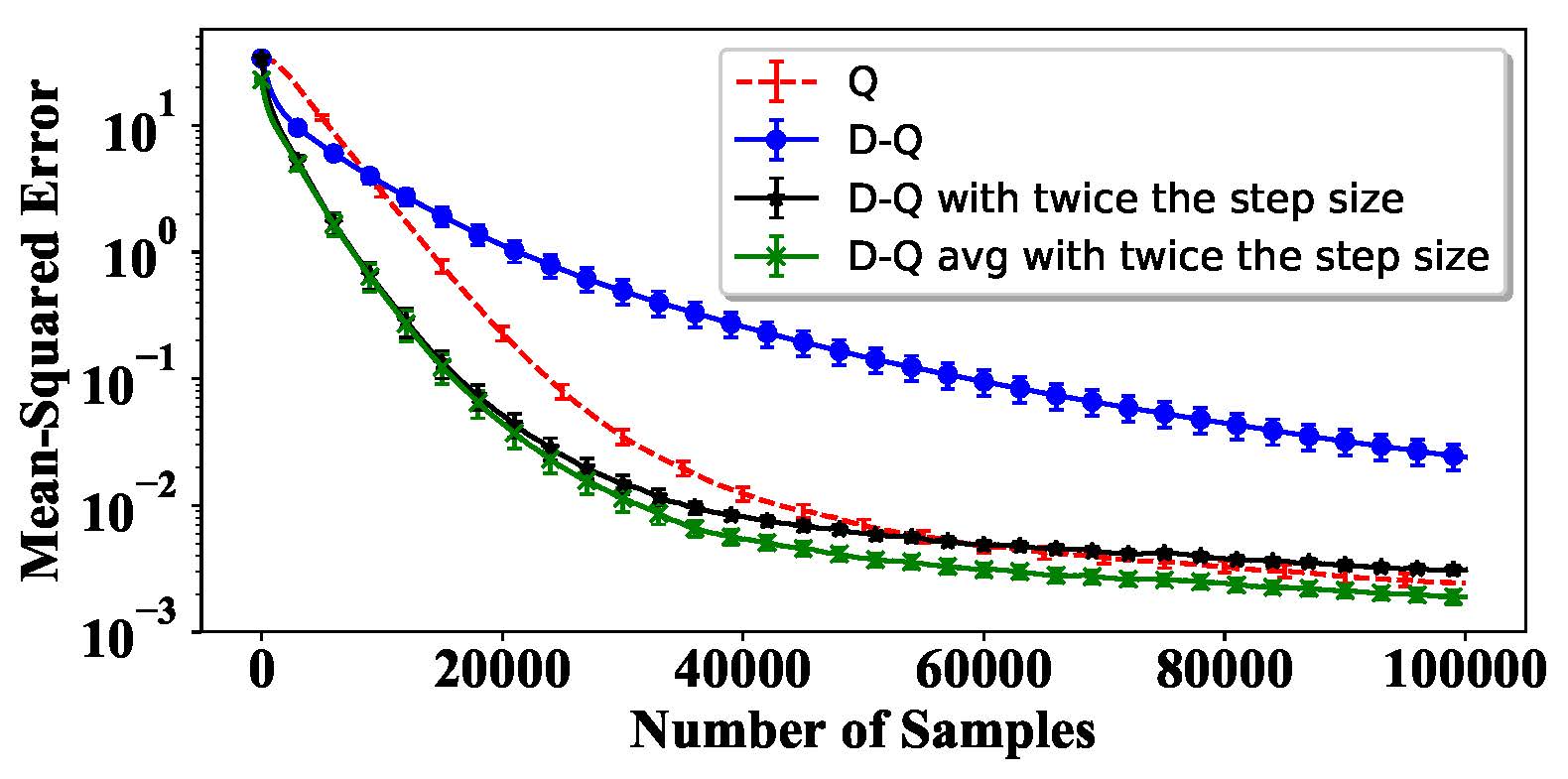}
        \caption{$4\times 4$ GridWorld}
        \label{fig:4x4-grid}
    \end{subfigure}
\quad
    \begin{subfigure}{0.6\textwidth}
        \centering
        \includegraphics[width=2.5in]{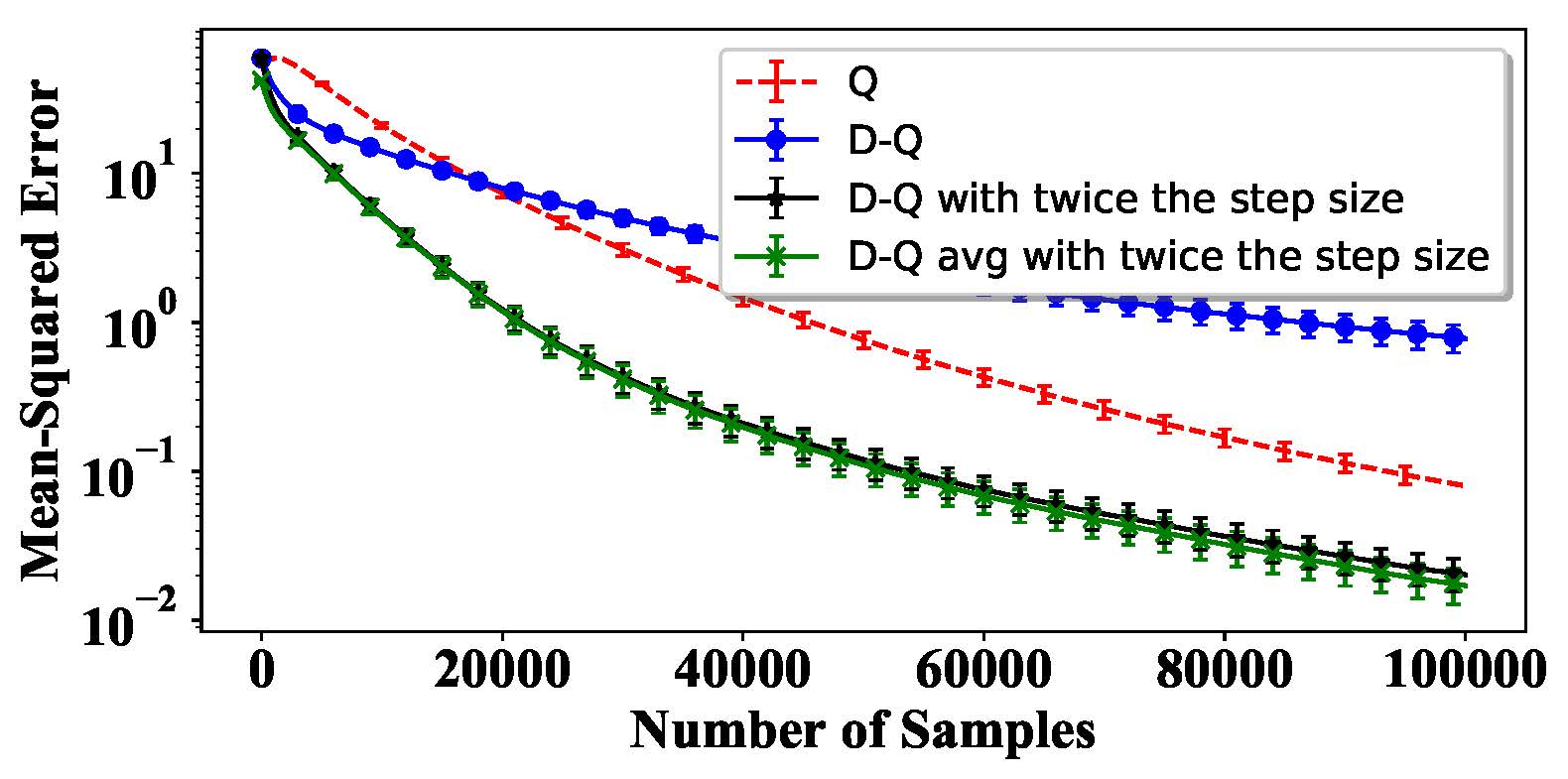}
        \caption{$5\times 5$ GridWorld}
        \label{fig:5x5-grid}
    \end{subfigure}
    \caption{Simulation results for GridWorld with dimensions $3,4,5$. In all the three simulations, Double Q-learning with twice the step-size and averaged output outperforms Q-learning.}
    \label{fig:gridWorld-result}
\end{figure}

The second environment we simulate is the GridWorld game with a similar setting as in \cite{geramifard_2013}. Consider a $n\times n$ grid where the agent starts at position $(1,1)$ and the goal is to reach the position $(n,n)$. A $3\times 3$ GridWorld is shown in Fig. \ref{fig:grid-Example}.
For each step, the agent can walk in four directions: up, down, left or right. If the agent walks out of the grid, the agent will stay at the same cell. There is a 30$\%$ probability that the chosen direction is substituted by any one of the four directions randomly. The agent receives reward $-10^{-3}$ in each step, but receives reward $1$ at the destination. The game ends when the agent arrives at the destination. We consider GridWorld with $n = 3,4$ and $5$, so the number of pairs of states and actions can be up to $100$. The discount factor is set as $\gamma = 0.9$. We run tabular Q-learning and tabular Double Q-learning.  Simulation results are shown in Fig. \ref{fig:gridWorld-result}.

 As we can see, Double Q-learning using step size $\alpha_n$ converges much slower than all the other three algorithms even though it has a slightly better asymptotic variance as shown in Fig. \ref{fig:3x3-grid}. By simply doubling the step-size and using the averaged output, Double Q-learning outperforms Q-learning in all the three settings. It is worth pointing out that theoretically speaking, Theorem \ref{thm: comparison} does not apply to this example because the optimal policy is not unique. However, the insights offered by Theorem \ref{thm: comparison} still hold.

\subsection{CartPole}
The third experiment we conduct is the classical CartPole control problem introduced in \cite{barto_1983}. In this problem, a cart with a pole is controlled by applying a force, either to left or to right. The goal is to keep the pole upright for as long as possible. The player receives a $+1$ reward for every time step until the episode ends which happens when the pole falls down or the cart moves out of a certain region.  Unlike the previous numerical results which mainly focus on the mean-squared error, in this case, we study how fast the four algorithms can find a policy that achieves the best performance. We train algorithms on CartPole-v0 available in OpenAI Gym \cite{brockman_2016}. Specifically, we consider Q-learning and Double Q-learning equipped with $\epsilon$-greedy exploration. The training is episodic, in the sense that for each episode, i.e., the step-size and the $\epsilon$ are updated after one episode. In particular, for the $n$th episode, we use
$
\epsilon_n = \max(0.1,\min(1,1-\log(\frac{n}{200}))), \alpha_n = \frac{40}{n+100}.
$
The step size is different from previous experiments because we only train $1000$ episodes for CartPole, and therefore, the step-size would have remained too large throughout if we had used the previous step-size rule and we noticed that this leads to convergence issues. 
The discount factor is set as $\gamma = 0.999$. Since the state space of CartPole is continuous, we discrete it into $72$ states following \cite{CartPole_Blog}.

\begin{figure}[!ht]
\begin{minipage}[h]{0.4\textwidth}
\includegraphics[width=2.5in]{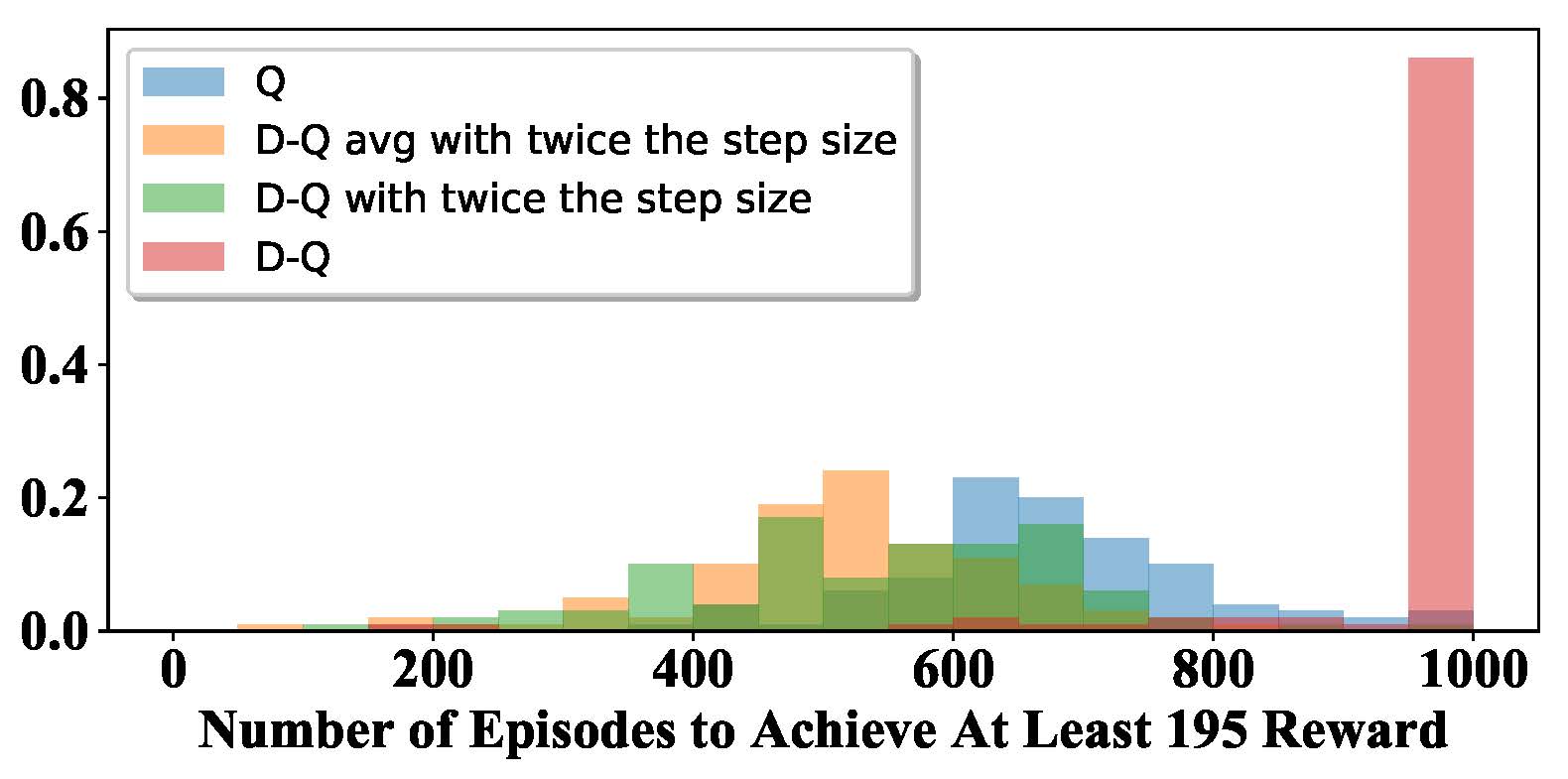}
\end{minipage}
\qquad
\begin{minipage}[t]{0.3\textwidth}
\begin{tabular}[c]{|l|l|}
\hline
Algorithm                      & Mean Hit Time     \\ \hline
Q                              & $645.0 \pm 12.93$ \\
D-Q avg with twice the step size & $487.5 \pm 12.19$ \\
D-Q with twice the step size  & $518.0 \pm 14.77$ \\\hline
\end{tabular}
\end{minipage}
\captionlistentry[table]{A table beside a figure}
\captionsetup{labelformat=andtable}
\caption{Distribution of "hit time", i.e., number of episodes needed to obtain a mean reward of 195 in CartPole-v0, with the number of episodes capped at $1000$. The mean hit time of each algorithm is summarized with its standard deviation.}
\label{fig:hit-time-cartpole}
\end{figure}

We evaluate the algorithms based on their "hit time", i.e., the time at which they first learn a fairly good policy. We say an algorithm learns a fairly good policy if the mean reward of the greedy policy based on the estimator learned from the first $n$ episodes exceed $195$.
To reduce the computational overhead, we evaluate the policy obtained after every $50$ episodes by averaging the reward obtained by the policy over $1000$ independently run episodes. The distribution of the "hit time" for each algorithm in $100$ independent tests is shown in Fig. \ref{fig:hit-time-cartpole}. We observe that Double Q-learning using the same learning rate performs much worse than other algorithms. However, when using twice the step size, Double Q-learning finds a good policy faster than Q-learning, at the cost of a larger standard deviation for the "hit time". The increase of variance can be mitigated by using the averaged estimator, which at the same time improves the convergence speed. 
\setlength{\belowcaptionskip}{-5pt}

\vspace{-0.25cm}
\subsection{Maximization Bias of Q-learning}
The fourth example we investigate is the maximization bias example similar to that in \cite[Page 135]{Sutton_18}. Since Double Q-learning was proposed to alleviate the maximization bias from Q-learning. we study how the proposed modification, doubling the step size and averaging the two estimators in Double Q-learning, affects the performance in an example where Double Q-learning is known to be helpful. To be specific, there are $M+1$ states labelled as $\{0,\cdots,M\}$ with two actions, left and right. The agent starts at state $0$. If the agent goes to the right, the game ends, but if she moves to the left, she goes with equal probability to one of the other $M$ states. Both actions result in zero reward. When the agent is at state $1$ to state $M$, if she goes to the right, she returns to state $0$; if she goes to the left, the game ends. Both actions result in a reward independently sampled from a normal distribution with mean $-0.1$ and standard deviation $1$. 

\begin{figure}[!ht]

    \begin{subfigure}{0.5\textwidth}
    \centering
\includegraphics[width=3in]{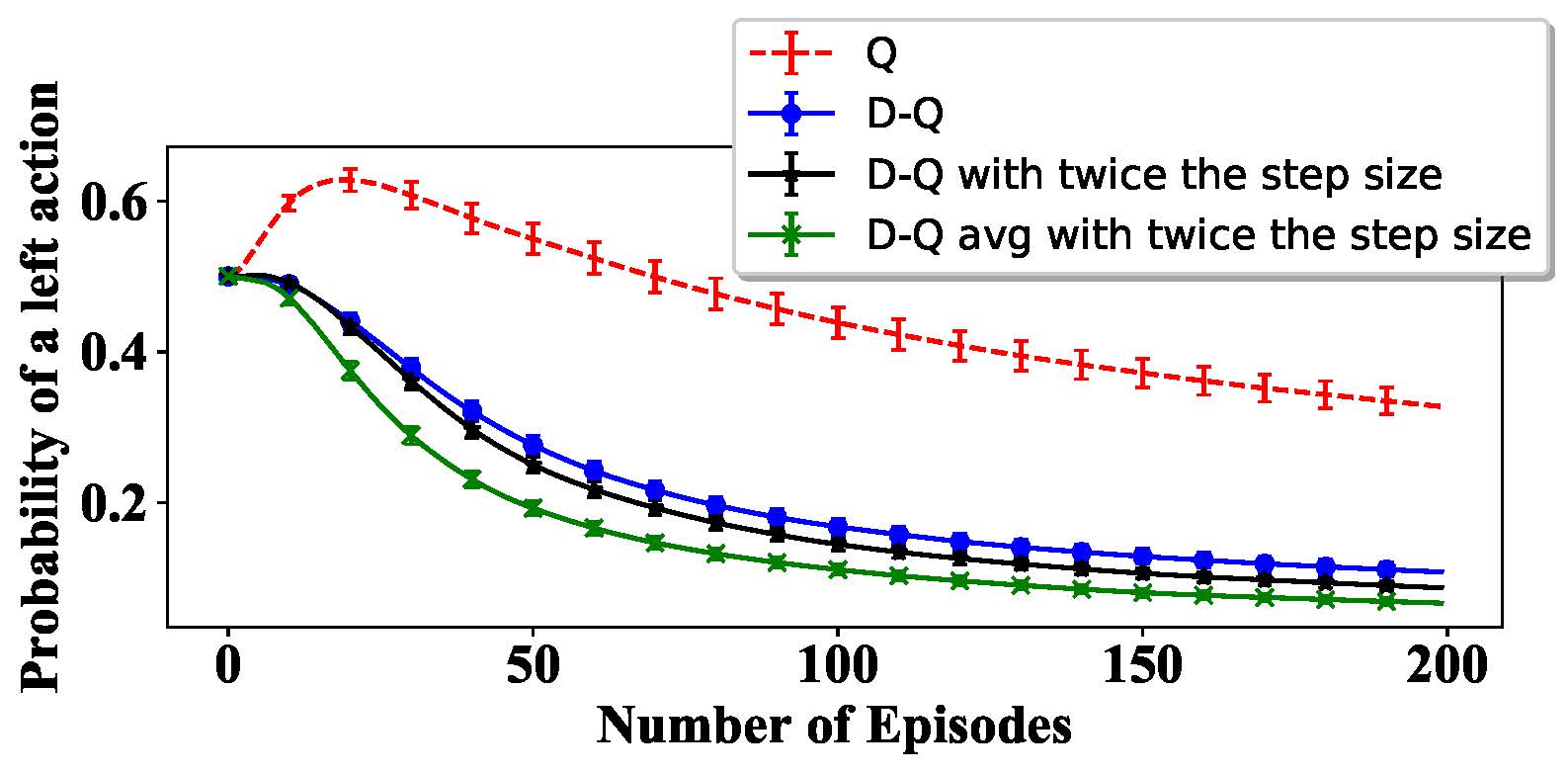}
    \caption{In a tabular setting with $M = 8$}
    \label{fig:bias-tabular}
    \end{subfigure}
    \begin{subfigure}{0.5\textwidth}
    \centering
\includegraphics[width=3in]{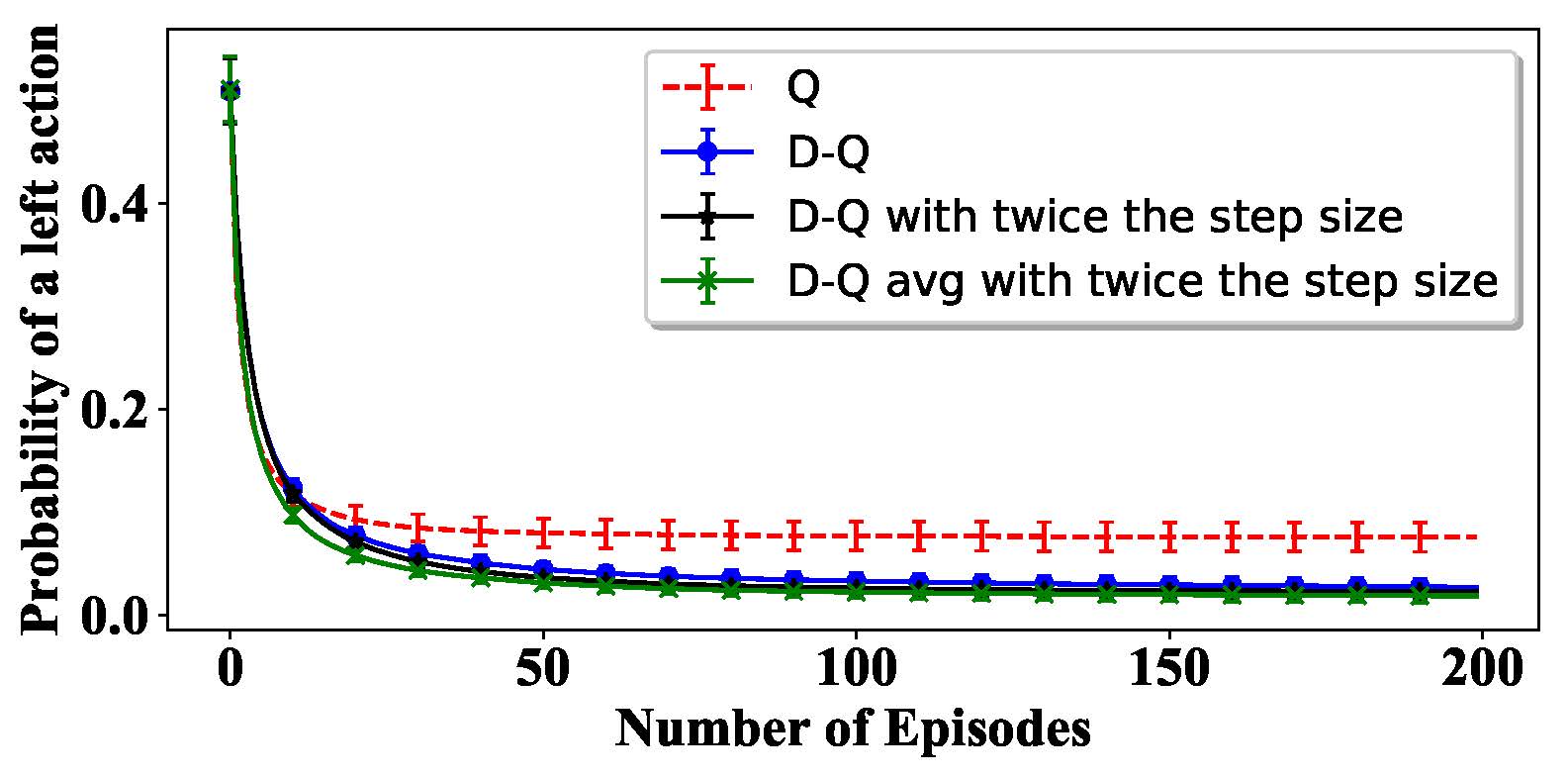}
    \caption{In a setting with neural network function approximations and $M=10^9$}
    \label{fig:bias-nn}
    \end{subfigure}
\qquad
\caption{The probability to go to the left for different algorithms in an environment similar to the maximization bias example from \cite{Sutton_18}. A lower probability indicates a better policy.}
\label{fig:bias-result}
\end{figure}

We first test the algorithms in a tabular setting with $M = 8$. The exploration policy is set to be $\epsilon$-greedy with $\epsilon = 0.1$. In the $n$th episode, $\alpha_n = \frac{10}{n+100}$. We train the algorithms for $200$ episodes. All estimators are initialized as zero. To evaluate the algorithms, we plot the probability of the agent going left after every episode. In particular, at the end of $n$ episodes, we count how often the estimated Q-function of a left action is larger than that of a right action at state $0$. In addition, the probability is taken to be the average of $1000$ independent runs. Notice that going right always maximizes the mean reward for the agent, so a larger probability to go left indicates that the algorithm has learned a worse policy. The result is shown in Fig. \ref{fig:bias-tabular}. As we can see, Q-learning suffers from the maximization bias when the number of episodes is small since there is a large probability of going to the left. On the other hand, there is no such problem with Double Q-learning. Furthermore, Double Q-learning with twice the step size and averaging improves performance even more.

In addition to the tabular setting, we also explore a setting where neural networks are used for function approximations. In particular, we consider the same environment as before, but with $M = 10^9$. In this way, it is infeasible to use a table for the whole $Q-$function. We assume that the $Q-$function is approximated by a neural network with two hidden layers of dimension $4$ and $8$. Each pair of adjacent layers is fully connected, with ReLU as the activation function. We use stochastic gradient descent with no momentum as the optimizer. Other settings are the same as those in the tabular setting. The result is shown in Fig. \ref{fig:bias-nn}. We can see that although Q-learning does not seem to suffer from maximization bias any more, it performs worse than Double Q-Learning. In addition, Double Q-Learning with twice the step size and averaging helps improve the performance.
\setlength{\belowcaptionskip}{-5pt}

\section{Conclusion}

It is known from prior work that Q-learning has faster convergence rate while Double Q-learning has better mean-squared error. A natural attempt to improve the convergence rate of Double Q-learning is to increase its stepsize (also called learning rate), but this leads to worse mean-squared error. We theoretically showed that increasing the learning rate of Double Q-learning while using a simple averaging at the output improves its convergence rate while making the mean-squared error equal to that of Q-learning. In the supplementary material, we further expand on our theoretical results. Our theoretical results are further supported by numerical experiments which also provide some useful guidelines for implementations of Double Q-learning. However, these results do not immediately apply to Double Q-learning with nonlinear function approximation, which we leave for future investigation.

\section*{Broader Impact}

Reinforcement learning (RL) has been the driving force behind many recent breakthroughs in Artificial Intelligence, including defeating humans in games (e.g., chess, Go, StarCraft), self-driving cars, smart home automation, among many others. However, much of the successes build on efficient heuristics and  empirical explorations, lacking sufficient theoretical understanding. One such example is Double Q-learning, which is the common practice used in deep reinforcement learning. This work establishes a theoretical analysis of the mean-squared error of double Q-learning, and provides principled guidelines for its implementation. These contributions have the potential to promote a stronger understanding of common RL algorithms both in theory and practice, accelerate the design of more efficient, interpretable RL algorithms, and benefit tremendous RL-driven  applications that are societally impactful.
\paragraph{Acknowledgment:} The work of Wentao Weng was conducted during a visit to the Coordinated Science Lab, UIUC during 2020. Research is also supported in part by ONR Grant N00014-19-1-2566, NSF/USDA Grant AG 2018-67007-28379, ARO Grant W911NF-19-1-0379,  NSF Grant CCF 1934986.

\bibliographystyle{plain}
\bibliography{wentao} 
\newpage
\appendix

\section{Linearization Results}
In this section, we provide more details on the derivation of the results pertaining to the asymptotic mean-squared errors in Theorem \ref{thm: comparison}.  While \cite{Devraj_2017} provides an outline of the result, we provide some missing details here, including additional assumptions under which the result in \cite{Devraj_2017} is valid. The following result from \cite{Benveniste_12} will be useful to us. 
\subsection{Central Limit Theorem for SA}
Statements in this part are adapted from \cite[Chapter 2 and 3]{Benveniste_12}. Consider a SA algorithm of the form 
\begin{equation}
\xi_n = \xi_{n-1}+\gamma_n W(\xi_{n-1},Y_n),
\end{equation}
where $\xi_n$ lies in $\mathbb{R}^d$, and the state $Y_n$ lies in $\mathbb{R}^k$. Suppose the algorithm satisfies following assumptions.
\begin{assumption}\label{as:sa-algo}
\cite[Page 43, Assumption A]{Benveniste_12}

\textbf{(a). Decreasing Step Size:}
\begin{equation}
\gamma_n \geq 0;~\sum_{n}\gamma_n=+\infty;~\sum_n \gamma_n^{\alpha} < \infty\mspace{10mu}\text{for some }\alpha > 1.
\end{equation}
\textbf{(b). Markovian Noise:} There exists a Markov chain $\{\eta_n\},$ independent of $\{\xi_n\}$ with a unique stationary distribution such that $Y_n=f(\eta_n).$

\textbf{(c). Existence of a Mean Vector Field:} We assume the existence of the mean vector field defined by 
\[
w(\xi):=\lim_{n \to \infty} \expect{W(\xi,Y_n)},
\]
where the expectation is taken under the distribution of $(Y_n).$ 
\end{assumption}
Assumption \ref{as:sa-algo}(c) allows us to introduce the ODE
\begin{equation}\label{def:ode}
\dot{\xi} = w(\xi), \xi(0) = z
\end{equation}
whose unique solution is denoted as $[\xi(z,t)]_{t \geq 0}.$
The next assumption we have is on the ODE. 
\begin{assumption}\label{as:attract-domain}\cite[Assumption (A.2), Assumption (A.2b)]{Benveniste_12}
The ODE (\ref{def:ode}) has an attractor $\xi^*$, whose domain of attraction is denoted by $D_*$. Assumption \ref{as:sa-algo} is satisfied in $D_*$.
\end{assumption}
Further, we assume the uniqueness of the attractor.
\begin{assumption}\label{as:sa-unique-point}\cite[Page 108]{Benveniste_12}
The ODE is globally asymptotically stable with a unique stable equilibrium point $\xi^*$.
\end{assumption}
Define 
\begin{equation}\label{eq:derivative}
C(\xi) := \sum_{n=-\infty}^{+\infty} \cov[W(\xi,Y_n), W(\xi,Y_1)]
\end{equation}
where $\cov$ denotes the covariance when $Y_1$ is stationary. We can now state the central limit theorem.
\begin{theorem}\label{thm:sa-clt}\cite[Page 110, Theorem 3]{Benveniste_12}
Suppose Assumption \ref{as:attract-domain} and Assumption \ref{as:sa-unique-point} hold, and the step size sequence satisfies $\gamma_n = \frac{1}{n}$. If $\nabla_\xi w(\xi^*)$ and $C(\xi^*)$ exist, and $\lambda_{\max}(\nabla_\xi w(\xi^*)) < -\frac{1}{2}$, we have
\begin{equation}
n^{\frac{1}{2}}(\xi_n-\xi^*) \xrightarrow[d]{} \mathcal{N}(0,P)
\end{equation}
where $P$ is the unique symmetric solution of the Lyapunov equation
\[
\left(\frac{I}{2}+\nabla_\xi w(\xi^*)\right)P+P\left(\frac{I}{2}+\transpose{\nabla_\xi w(\xi^*)}\right)+C(\xi^*) = 0.
\]
\end{theorem}
\subsection{Applications to Q-learning and Double Q-learning}
In this section, we show that Theorem \ref{thm:sa-clt} is applicable to Q-learning (\ref{eq:Q-update}) and Double Q-learning (\ref{eq:doubleQ-update}) under the assumptions stated in the main body of the paper. Note that the step sizes are assumed to be $\alpha_n = \frac{g}{n}$, and $\delta_n = \frac{2g}{n}$ in Theorem \ref{thm: comparison}, which are different from that in Theorem \ref{thm:sa-clt}. Therefore, we scale the reward function and feature vectors to absorb the constant $g$ (or $2g$) in updates of Q-learning and Double Q-learning. The step sizes are then shifted to $\frac{1}{n}$.

Recall $Z_n = (X_n,S_{n+1})$ defined in the proof of Theorem \ref{thm: comparison}. We first notice that Assumption \ref{as:sa-algo} is automatically satisfied because: 1) The step size condition is fulfilled for $\frac{1}{n}$; 2) The samples $\{Z_n, n \geq 0\}$ form a Markov chain independent of $\theta_n$; 3) The mean vector field $w(\theta)$ is well-defined since $\{Z_n\}$ has a unique limiting stationary distribution, and its state space $\mathcal{X} \times \mathcal{S}$ is finite. As a result, the ODE for Q-learning is defined as 
\begin{equation}\label{eq:q-ode}
\dot{\theta}(t) = g\expect{\phi(X_n)(R(X_n)+\gamma H(\theta(t),\theta(t),S_{n+1})-\transpose{\phi(X_n)}\theta(t))},
\end{equation}
and that of Double Q-learning is given by
\begin{subequations}\label{eq:doubleq-ode}
\begin{align}
\dot{\theta}^A(t) &= g\expect{\phi(X_n)(R(X_n)+\gamma H(\theta^A(t),\theta^B(t),S_{n+1})-\transpose{\phi(X_n)}\theta^A(t))}, \\
\dot{\theta}^B(t) &= g\expect{\phi(X_n)(R(X_n)+\gamma H(\theta^B(t),\theta^A(t),S_{n+1})-\transpose{\phi(X_n)}\theta^B(t))}.
\end{align}
\end{subequations}
For ease of notation, denote $U(t) = ((\theta^A(t));(\theta^B(t)))$. The notation $(\bold{a};\bold{b})$ is a vector that is the concatenation of $\bold{a}$ and $\bold{b}$. Also, denote the right hand side of (\ref{eq:q-ode}) by $w(\theta(t))$, and that of (\ref{eq:doubleq-ode}) by $\tilde{w}(U(t))$.

To guarantee Assumption \ref{as:attract-domain} and Assumption \ref{as:sa-unique-point}, we make the following assumption.
\begin{assumption}\label{as:ode-gas}
Both $\theta(t)$ and $U(t)$ have unique globally asymptotically stable (GAS) equilibrium points. 
\end{assumption}  
Sufficient conditions under which Q-learning with linear function approximation satisfies Assumption \ref{as:ode-gas} are studied in \cite{Lee_Dong_He_2019, Melo_2008}. While little is known on the convergence of Double Q-learning with linear function approximation, it is commonly perceived that double Q-learning is more stable than Q-learning even when equipped with neural networks \cite{Hasselt_16}. 

Denote the unique stable point of $\theta(t)$ as $\theta^*$, and that of $U(t)$ as $U^*$. It is shown in \cite{Melo_2008} that $\theta^*$ is the solution to the projected Bellman equation. The following lemma shows that $(\theta^*;\theta^*)$ is also the GAS equilibrium point of the ODE of Double Q-learning. The reader is referred to the next section for the proof.
\begin{lemma}\label{lemma:q-ode-fix}
The point $U^*$ is exactly $(\theta^*;\theta^*)$.
\end{lemma}

To apply Theorem \ref{thm:sa-clt}, we need to work out $\nabla_{\theta}w(\theta^*), C_{\theta}(\theta^*), \nabla_U\tilde{w}(U^*), C_U(U^*)$ which are the analogs of the quantities in (\ref{eq:derivative}) for Q-learning and Double Q-learning, respectively. However, since the function $H$ in (\ref{eq:q-ode}) could be non-differentiable around $\theta^*$, we impose the following assumption from \cite{Devraj_2017} that ensures the existence of $\nabla_{\theta}w(\theta^*)$ and $ \nabla_U\tilde{w}(U^*)$.
\begin{assumption}\label{as:unique-optimal}
The optimal policy $\pi^*:=\pi_{\theta^*}$ is unique.
\end{assumption}
Under this assumption, we summarize the exact forms of $\nabla_{\theta}w(\theta^*), C_{\theta}(\theta^*), \nabla_{U}\tilde{w}(U^*), C_U(U^*)$ in the following result. 
The proof of this lemma is deferred to the next section.
\begin{lemma}\label{lemma:exact-linear}
Following the notation in the proof of Theorem \ref{thm: comparison}, the following equalities hold:
\begin{subequations}\label{eq:q-doubleq-ode}
\begin{align}
\nabla_{\theta}w(\theta^*) &= g\bar{A}, C_{\theta}(\theta^*) = g^2(B_1 + B_2); \\
\nabla_{U}\tilde{w}(U^*) &= g\bar{A}_D, C_{U}(U^*) = 2g^2\begin{pmatrix}B_1 & B_2 \\ B_2 & B1\end{pmatrix},
\end{align}
where $B_2 := \frac{1}{2}\expect{\sum_{n=2}^{\infty} (W(Z_n)\transpose{W(Z_1)} + W(Z_1)\transpose{W(Z_n)})}$, $B_1 := \expect{W(Z_1)\transpose{W(Z_1)}} +  B_2$, and 
$
W(Z_n) := \left(b(Z_n)+A_2(Z_n)\theta^* - A_1(Z_n)\theta^*\right).$
\end{subequations}
\end{lemma}
Note that in Theorem \ref{thm: comparison}, we assume $\theta^* = 0$. Therefore, $W(Z_n) = b(Z_n).$

Define $g_0:=\inf\{g\geq 0: g\max(\lambda_{\max}(\bar{A}),\lambda_{\max}(\bar{A}_D)) < -1\}$. Then whenever $g > g_0$, we have $\lambda_{\max}(\nabla_{\theta}w(\theta^*)) < -\frac{1}{2}, \lambda_{\max}(\nabla_U \tilde{w}(U^*)) < -\frac{1}{2}$. So far we have checked all conditions in Theorem \ref{thm:sa-clt} for Q-learning and Double Q-learning. Therefore, the central limit theorem holds:
\begin{subequations}\label{eq:q-dq-clt}
\begin{align}
n^{\frac{1}{2}}(\theta_n - \theta^*) &\xrightarrow[d]{}\mathcal{N}(0,P_Q) \\
n^{\frac{1}{2}}(U_n - U^*) &\xrightarrow[d]{}\mathcal{N}(0,P_D) 
\end{align}
\end{subequations}
where $P_Q,P_D$ are given by
\begin{subequations}
\begin{align}
\left(\frac{I}{2}+g\bar{A}\right)P_Q+P_Q\left(\frac{I}{2}+g\transpose{\bar{A}}\right)+ g^2(B_1+B_2) &= 0 \label{eq:q-clt-lyap}\\
\left(\frac{I}{2}+g\bar{A}_D\right)P_D+P_D\left(\frac{I}{2}+g\transpose{\bar{A}}_D\right)+ 2g^2\begin{pmatrix}B_1 & B_2 \\ B_2 & B_1\end{pmatrix} &= 0\label{eq:dq-clt-lyap}.
\end{align}
\end{subequations}
We can see Eq. (\ref{eq:q-clt-lyap}) and Eq. (\ref{eq:dq-clt-lyap}) are indeed identical to the two equations, Eq. (\ref{eq:Q-LSA-result}) and Eq. (\ref{eq:doubleQ-LSA-result}), for the asymptotic covariance matrices of Q-learning and Double Q-learning. However, since we only establish convergence in distribution of a sequence of random vectors, it does not immediately imply that the limit of variances of these random vectors converges to the variance of the corresponding normal distribution. To fix this gap, we first observe that the function $\transpose{\bold{x}}\bold{x}$ is continuous where $\bold{x}$ is a vector. By the Continuous Mapping Theorem for random vectors and Eq. (\ref{eq:q-dq-clt}), it holds
\begin{subequations} \label{eq:norm-converge}
\begin{align}
n\norm{\theta_n - \theta^*}{2}^2 &\xrightarrow[d]{} \norm{\bold{X}_Q}{2}^2 \\
n\norm{(U_n - U^*)}{2}^2 &\xrightarrow[d]{} \norm{\bold{X}_D}{2}^2. 
\end{align}
\end{subequations}
where $\bold{X}_Q$ follows the normal distribution $\mathcal{N}(0,P_Q)$, and $\bold{X}_D$ follows $\mathcal{N}(0,P_D)$. Here, the convergence in distribution is for random variables. Finally, to establish the convergence of the mean of these random variables, we need uniform integrability, which we assume as follows. 
\begin{assumption}\label{as:q-square-converge}
The three sequences of random variables 
\[
\{n\norm{\theta_n-\theta^*}{2}^2, n \geq 1\}, \{n\norm{\theta_n^A-\theta^*}{2}^2, n \geq 1\}, \{n\norm{\theta_n^B-\theta^*}{2}^2, n \geq 1\}
\]
are all uniformly integrable.
\end{assumption}
 Assumption \ref{as:q-square-converge} directly implies the sequence $\{n\norm{U_n-U^*}{2}^2, n \geq 1\}$ is uniformly integrable. Combining (\ref{eq:norm-converge}) with Assumption \ref{as:q-square-converge}, we have 
 \begin{subequations}
 \begin{align}
 \lim_{n \to \infty} n\expect{\norm{\theta_n - \theta^*}{2}^2} &= \expect{\norm{\bold{X}_Q}{2}^2} = \trace(P_Q) \\
\lim_{n \to \infty} n\expect{\norm{(U_n - U^*)}{2}^2} &= \expect{\norm{\bold{X}_D}{2}^2} = \trace(P_D). 
 \end{align}
\end{subequations}
Under all the assumptions stated in this section, the linearizations in Section \ref{sec:linearization} are valid.

\subsection{Proof of Lemmas}
In this section, we provide missing proofs of Lemma \ref{lemma:q-ode-fix} and Lemma \ref{lemma:exact-linear}.

\paragraph{Proof of Lemma \ref{lemma:q-ode-fix}:}
By Assumption \ref{as:ode-gas}, the ODE of Double Q-learning has a unique GAS equilibrium point. Denote this point as $(\theta_1;\theta_2)$. By the symmetry of the ODE (\ref{eq:doubleq-ode}), $(\theta_2;\theta_1)$ is also a GAS equilibrium point of the ODE. But such point is unique. We thus have $\theta_1 = \theta_2$. In this case, the ODE (\ref{eq:doubleq-ode}) degenerates to the ODE (\ref{eq:q-ode}) of Q-learning. Therefore, we have $\theta_1 = \theta_2 = \theta^*$. \qed

\paragraph{Proof of Lemma \ref{lemma:exact-linear}:}
We show it for Q-learning. The same strategy can be applied to Double Q-learning. 

Recall the ODE of Q-learning defined as (\ref{eq:q-ode}). We know that $\theta^*$ is the unique GAS equilibrium point of this ODE. Recall that the right hand side of (\ref{eq:q-ode}) is denoted by $w(\theta(t))$. Then at the point $\theta^*$, the following equality holds:
\[
w(\theta^*) = g\left(\expect{\phi(X_n)R(X_n)}+\gamma \expect{\phi(X_n) H(\theta^*,\theta^*,S_{n+1})} - \expect{\phi(X_n)\transpose{\phi(X_n)}}\theta^*\right).
\]
Note that the optimal policy $\pi^*$ is unique by assumption. We can rewrite $H(\theta^*,\theta^*,S_{n+1})$ as $\transpose{\phi(S_{n+1},\pi^*(S_{n+1}))}\theta^*$. Then we can see 
\begin{align}
w(\theta^*) &= g\left(\expect{\phi(X_n)R(X_n)}+\gamma \expect{\phi(X_n)\transpose{\phi(S_{n+1},\pi^*(S_{n+1}))}\theta^*}-\expect{\phi(X_n)\transpose{\phi(X_n)}}\theta^*\right) \\
&= g\expect{\phi(X_n)R(X_n)}+g( \bar{A}_2-\bar{A}_1)\theta^*.
\end{align}
which is the same as the ODE of the linearization (\ref{eq:Q-update linear}) at the point $\theta^*$.

Furthermore, since the optimal policy is unique for $\theta^*$, we can define a constant \[\omega \coloneqq \min_{(s,a) \in \mathcal{X} \colon a \not = \pi^*(s)} (\transpose{\phi(s,\pi^*(s))}\theta^* - \transpose{\phi(s,a)}\theta^*) > 0\] be the minimum gap between value functions of optimal actions and non-optimal actions for all states, estimated by $\theta^*$. Let $\epsilon = \frac{\omega}{3\norm{\Phi}{1}}$. Consider any $\theta \in \mathbb{R}^{d} $ satisfying $\norm{\theta-\theta^*}{\infty} \leq \epsilon$. We claim that the greedy policy $\pi_{\theta}$ is equal to $\pi^*$. To see that it is true, let us fix a state $s \in \mathcal{S}$. For any $a \in \mathcal{A}$ and $a \not = \pi^*(s)$, it holds
\[
\transpose{\phi(s,\pi^*(s))}\theta_a - \transpose{\phi(s,a)}\theta_a \geq \transpose{\phi(s,\pi^*(s))}\theta^* - \transpose{\phi(s,a)}\theta^* - 2\norm{\transpose{\Phi}(\theta-\theta^*)}{\infty} \geq \omega - \frac{2\omega}{3} > 0.
\]
Therefore, $\pi_{\theta} = \pi^*.$ Consequently, for any $\theta$ such that $\norm{\theta-\theta^*}{\infty} \leq \epsilon$, it holds 
$w(\theta) = g\expect{\phi(X_n)R(X_n)}+g( \bar{A}_2-\bar{A}_1)\theta$. Therefore, $\nabla_{\theta}w(\theta^*) = g\widebar{A}=g(\bar{A}_2-\bar{A}_1).$

For $C_{\theta}(\theta^*)$, define 
\[
W(Z_n) := \phi(X_n)R(X_n)+\gamma \phi(S_{n+1},\pi^*(S_{n+1}))\theta^* - \phi(X_n)\transpose{\phi(X_n)}\theta^*.
\]
Then by definition,
\[
\begin{aligned}
C_{\theta}(\theta^*) &= \sum_{n=-\infty}^{+\infty}  \expect{(gW(Z_n) - w(\theta^*))\transpose{(gW(Z_1) - w(\theta^*))}} \\
&= g^2\sum_{n=-\infty}^{+\infty} \expect{(W(Z_n))\transpose{(W(Z_1))}} \\
&= g^2\left(\expect{(W(Z_1))\transpose{(W(Z_1))}}+\sum_{n=2}^{+\infty} \expect{W(Z_1)\transpose{W(Z_n)} + W(Z_n)\transpose{W(Z_1)}}\right).
\end{aligned}
\]
\qed

\section{A Stronger Result for the Mean-Squared Error}
In this section, we provide a stronger result for the asymptotic mean-squared error of Double Q-learning. Assume that the vector $b(x)$ defined in the proof of Theorem \ref{thm: comparison} is not the same for all $x \in \mathcal{X}$. Additionally, assume that $\theta^* = 0$. Following the notation in Theorem \ref{thm: comparison}, we have this result.
\begin{theorem}\label{thm:worse-error}
Let the step sizes of Q-learning and Double Q-learning be $\alpha_n=g/n$ and $\delta_n=2g/n$ respectively, where $g$ is a positive constant. With the same constant $g_0$ in Theorem \ref{thm: comparison}, for any $g > g_0$,  it holds
\[
\amse(\theta^A) \geq \amse(\theta) + c_0g
\]
where $c_0$ is a positive constant independent from $g$.
\end{theorem}

Theorem \ref{thm:worse-error} shows that in general, the asymptotic mean-squared error of Double Q-learning is worse than that of Q-learning, when using twice of the step size. Moreover, the gap scales at least linearly with respect to the step size. 

To prove Theorem \ref{thm:worse-error}, we need two additional lemmas. The first lemma is on the relationship between the two matrices $\bar{A}_D$ and $\bar{A}$ defined in the proof of Theorem \ref{thm: comparison}. 
\begin{lemma}\label{lemma:eigen-relation}
Following the notation in the proof of Theorem \ref{thm: comparison}, consider the matrix $\bar{A}_D = \begin{pmatrix} -\bar{A}_1 & \bar{A}_2 \\ \bar{A}_2 & -\bar{A}_1\end{pmatrix}$. The set of its eigenvalues is given by the union of eigenvalues of $\widebar{A}_2-\widebar{A}_1$ and that of $-(\widebar{A}_2+\widebar{A}_1)$.
\end{lemma}

\paragraph{Proof of Lemma~\ref{lemma:eigen-relation}:}
Suppose $\lambda$ is an eigenvalue of $\bar{A}_D$ with an eigenvector $v = \transpose{(\transpose{v_1},\transpose{v_2})} \not = 0$ where $v_1,v_2 \in \mathbb{R}^{d}$. We claim that $\lambda$ is either an eigenvalue of $-\bar{A}_1+\bar{A}_2$ or an eigenvalue of $-(\bar{A}_1+\bar{A}_2)$. To see this fact, it holds
\[
\bar{A}_D\begin{bmatrix} v_1 \\ v_2 \end{bmatrix} = \lambda \begin{bmatrix} v_1 \\ v_2 \end{bmatrix}.
\]
If $v_1+v_2 \not = \bold{0}$, then
\[
(-\bar{A}_1+\bar{A}_2)(v_1+v_2) = \lambda (v_1+v_2),
\]
showing that $\lambda$ is an eigenvalue of $-\bar{A}_1+\bar{A}_2$. Otherwise, suppose $v_1 + v_2 = \bold{0}$. Then $v_1 = -v_2$, and 
\[
-(\bar{A}_1+\bar{A}_2)v_1=\lambda v_1.
\]
We can also show that for every eigenvalue of $-\bar{A}_1+\bar{A}_2$ and $-(\bar{A}_1+\bar{A}_2)$, we can construct a corresponding eigenvector with respect to $\bar{A}_D$. Therefore, the set of eigenvalues of $\bar{A}_D$ is exactly the union of eigenvalues of $-\bar{A}_1+\bar{A}_2$ and $-(\bar{A}_1+\bar{A}_2)$. \qed

The second lemma is on the trace of the solution of a Lyapunov equation. 
\begin{lemma}\label{lemma:lyap-psd}
Consider a Lyapunov equation 
\[
AX+X\transpose{A}+Q=0,
\]
where $A,Q \in \mathbb{R}^{n\times n}$ are given, for some positive integer $n$. If $A$ is Hurwitz, and $Q\succcurlyeq 0$, and $\trace(Q) > 0$, then $\trace(X) > 0.$
\end{lemma}
Note that the notation $Q \succcurlyeq 0$ means that $Q$ is a positive semi-definite matrix.
\paragraph{Proof of Lemma~\ref{lemma:lyap-psd}:}
By \cite[Theorem 5.6]{Chen_1998}, if $A$ is Hurwitz, then $X$ has a unique solution that can be expressed as 
\begin{equation}\label{eq:solution-lyap}
X = \int_0^{\infty} \! e^{At}Qe^{\transpose{A}t} \, \mathrm{d}t.
\end{equation}
Since $Q \succcurlyeq 0$ by assumption, and $\transpose{(e^{At})} = e^{\transpose{A}t}$ for all $t$, we have $X \succcurlyeq 0$. We prove $\trace(X) > 0$ by contradiction. Suppose $\trace(X) = 0$. Therefore, as $X \succcurlyeq 0$, we have: $\transpose{\bold{v}}X\bold{v} = 0, \forall \text{ vectors } \bold{v}$ (since all eigenvalues of $X$ are $0$).

Denote the largest eigenvalue of $Q$ as $\lambda_m$, which must be a positive real value because $Q \succcurlyeq 0$ and $\trace(Q) > 0$. Suppose $\bold{v}$ is the unit eigenvector corresponding to $\lambda_m$, i.e., $Q\bold{v} = \lambda_m \bold{v}$, and $\norm{\bold{v}}{2} = 1.$ We have
\begin{equation}\label{eq:lemma2-intproduct}
\transpose{\bold{v}}X\bold{v} = \int_{0}^{\infty} \!\transpose{\bold{v}}e^{At}Qe^{\transpose{A}t} \bold{v}\,\mathrm{d}t.
\end{equation}
Note that $\lim_{t \to 0} e^{At} = I$, and $\lim_{t \to 0} e^{\transpose{A}t} = I.$ Therefore, for $\epsilon = \min\left(\frac{\lambda_m}{\norm{Q}{2}},1\right)$, there exists a $\tilde{t} > 0$, such that for any $0 \leq t \leq \tilde{t}$, we have
\begin{equation}\label{eq:lemma2-bound-norm}
\norm{e^{At}-I}{2} \leq \epsilon,\mspace{23mu}\norm{e^{\transpose{A}t}-I}{2} \leq \epsilon.
\end{equation}
Equation (\ref{eq:lemma2-intproduct}) can be rewritten as 
\begin{align}
\transpose{\bold{v}}X\bold{v} &= \int_0^{\tilde{t}} \! \transpose{\bold{v}}e^{At}Qe^{\transpose{A}t}\bold{v}\,\mathrm{d}t+\int_{\tilde{t}}^{\infty} \! \transpose{\bold{v}}e^{At}Qe^{\transpose{A}t}\bold{v}\,\mathrm{d}t \\
&\overset{a}{\geq} \int_0^{\tilde{t}} \! \transpose{\bold{v}}e^{At}Qe^{\transpose{A}t}\bold{v}\,\mathrm{d}t \\
&= \int_0^{\tilde{t}} \! \transpose{\bold{v}}(I+e^{At}-I)Q(I+e^{\transpose{A}t}-I)\bold{v}\,\mathrm{d}t \\
&= \int_0^{\tilde{t}} \! \transpose{\bold{v}}Q\bold{v}\,\mathrm{d}t+\int_0^{\tilde{t}} \! \transpose{\bold{v}}(e^{At}-I)Q\bold{v}\,\mathrm{d}t +  \int_0^{\tilde{t}} \! \transpose{\bold{v}}Q(e^{\transpose{A}t}-I)\bold{v}\,\mathrm{d}t \nonumber\\
&\mspace{23mu}+ \int_0^{\tilde{t}} \! \transpose{\bold{v}}(e^{At}-I)Q(e^{\transpose{A}t}-I)\bold{v}\,\mathrm{d}t. \label{eq:lemma2:decompose}
\end{align}
Inequality $a$ follows from the fact that $e^{At}Qe^{\transpose{A}t} \succcurlyeq 0$, for any $t \geq 0$. To lower bound (\ref{eq:lemma2:decompose}), we first have $\int_0^{\tilde{t}} \! \transpose{\bold{v}}Q\bold{v}\,\mathrm{d}t = \tilde{t}\norm{v}{2}^2\lambda_m$, by definition of $\bold{v}$. For the last three terms, using the definition of matrix norm and (\ref{eq:lemma2-bound-norm}), the following hold
\begin{align}
\left|\int_0^{\tilde{t}} \! \transpose{\bold{v}}(e^{At}-I)Q\bold{v}\,\mathrm{d}t\right| &\leq \tilde{t}\norm{v}{2}^2 \norm{Q}{2} \epsilon \\
\left|\int_0^{\tilde{t}} \! \transpose{\bold{v}}Q(e^{\transpose{A}t}-I)\bold{v}\,\mathrm{d}t\right| &\leq \tilde{t}\norm{v}{2}^2 \norm{Q}{2} \epsilon \\
\left|\int_0^{\tilde{t}} \! \transpose{\bold{v}}(e^{At}-I)Q(e^{\transpose{A}t}-I)\bold{v}\,\mathrm{d}t\right| &\leq \tilde{t}\norm{v}{2}^2 \norm{Q}{2} \epsilon^2.
\end{align}
Therefore, we have
\begin{equation}
\begin{aligned}
\transpose{\bold{v}}X\bold{v} &\geq \tilde{t}\norm{v}{2}^2\lambda_m - 2\tilde{t}\norm{v}{2}^2\norm{Q}{2}\epsilon - \tilde{t}\norm{v}{2}^2\norm{Q}{2}\epsilon^2 \\
&\geq \tilde{t}\norm{v}{2}^2 \left(\lambda_m - \norm{Q}{2}\left(2\epsilon+\epsilon^2\right)\right) \\
&\geq \frac{1}{2}\tilde{t}\norm{v}{2}^2\lambda_m
\end{aligned}
\end{equation}
by the definition of $\epsilon$. We can see that $\transpose{\bold{v}}X\bold{v} > 0$, which contradicts the assumption that $\transpose{\bold{v}}X\bold{v} = 0$. Therefore, $\trace(X) > 0$ by contradiction. \qed

We now present the proof of Theorem \ref{thm:worse-error}.
\paragraph{Proof of Theorem~\ref{thm:worse-error}:}
This proof follows the notation in the proof of Theorem \ref{thm: comparison}. In particular, we assume that the random vector $b(X_n)$ is centered at $0$. Recall Eq. (\ref{eq:analysis-equations}). Subtracting the block on the upper left corner by that on the upper right corner, we have
\begin{equation}\label{eq:gap-double}
(V-C)\transpose{\left(\frac{1}{2}I-g(\bar{A}_1+\bar{A}_2)\right)}+\left(\frac{1}{2}I-g(\bar{A}_1+\bar{A}_2)\right)(V-C)+2g^2(B_1 - B_2) = 0.
\end{equation}
By the definition of $B_1$ and $B_2$, we have $B_1 - B_2 = \expect{b(X_1)\transpose{b(X_1)}}$, whose trace is positive by assumptions. As in the proof of Theorem \ref{thm: comparison}, set the constant $g_0:=\inf\{g\geq 0: g\max(\lambda_{\max}(\bar{A}),\lambda_{\max}(\bar{A}_D)) < -1\}$. Since the matrix $\bar{A}_D$ is defined as $\begin{pmatrix} -\bar{A}_1 & \bar{A}_2 \\ \bar{A}_2 & -\bar{A_1}\end{pmatrix}$, we know by Lemma \ref{lemma:eigen-relation}, the set of eigenvalues of $-(\bar{A}_1+\bar{A}_2)$ is a subset of eigenvalues of $\bar{A}_D$. Therefore, for $g > g_0$, we have $g\lambda_{\max}(-(\bar{A}_1+\bar{A}_2)) < -1$. It immediately implies $\frac{1}{2}I - g(\bar{A}_1 + \bar{A}_2)$ is Hurwitz. Utilizing Lemma \ref{lemma:lyap-psd}, we have $\trace(V - C) > 0$. Together with the result $V + C = 2\Sigma_{\infty}^Q$ in the proof of Theorem \ref{thm: comparison}, we have 
\[
\amse(\theta^A) = \trace(V) = \trace(\Sigma_{\infty}^Q) + \frac{\trace(V-C)}{2} > \trace(\Sigma_{\infty}^Q) = \amse(\theta).
\]
On the other hand, to show $\amse(\theta^A) - \amse(\theta)$ indeed scales up linearly with respect to $g$, we divide both sides of Eq. (\ref{eq:gap-double}) by $g$
\[
(V-C)\transpose{\left(\frac{1}{2g}I-(\bar{A}_1+\bar{A}_2)\right)}+\left(\frac{1}{2g}I-(\bar{A}_1+\bar{A}_2)\right)(V-C)+2g(B_1 - B_2) = 0.
\]
Since $\frac{1}{2g}I-(\bar{A}_1+\bar{A}_2)$ is Hurwitz, the following equation has a unique positive definite solution $X$.
\[
X\transpose{\left(\frac{1}{2g}I-(\bar{A}_1+\bar{A}_2)\right)}+\left(\frac{1}{2g}I-(\bar{A}_1+\bar{A}_2)\right)X+(B_1 - B_2) = 0
\]
Therefore, $\trace(V-C) = 2g\trace(X)$.
Further, let $X'$ be the solution to the following Lyapunov equation
\[
X'\transpose{\left(-(\bar{A}_1+\bar{A}_2)\right)}+\left(-(\bar{A}_1+\bar{A}_2)\right)X'+(B_1 - B_2) = 0.
\]
Since $-(\bar{A}_1+\bar{A}_2)$ is Hurwitz, and $B_1 - B_2$ has a positive trace, we have $\trace(X') > 0$, which is independent of $g$. By the expression Eq. (\ref{eq:solution-lyap}) of $X$ and $X'$, it can be easily shown that $\trace(X) \geq \trace(X')$.
This proves that $\amse(\theta^A) - \amse(\theta) \geq c_0g$ for some positive constant $c_0$ independent from $g$. \qed

\end{document}